\documentclass[runningheads]{llncs}
\usepackage{graphicx}
\usepackage{comment} 
\usepackage{amsmath,amssymb} 
\usepackage{color}

\usepackage[table]{xcolor}
\usepackage{tabularx} 
\usepackage{multirow}
\usepackage{xfrac}
\usepackage{xspace} 
\usepackage{tikz}
\usepackage{xcolor}

\usetikzlibrary{svg.path}

\definecolor{boxGreen}{rgb}{0.44,0.68,0.28}
\definecolor{boxRed}{rgb}{0.75,0.,0.}
\definecolor{boxYellow}{rgb}{1.,0.75,0.}
\definecolor{boxBlue}{rgb}{0.27,0.45,0.77}
\definecolor{boxOrange}{rgb}{0.93,0.49,0.19}
\newcommand{\snippetBox}[2]{\draw[fill=#1,line width=0.5pt] (#2ex,0) rectangle +(1.25ex,1.25ex);}

\newcommand*{\eg}{e.g.\@\xspace}
\newcommand*{\ie}{i.e.\@\xspace}
\makeatletter
\newcommand*{\etc}{%
  \@ifnextchar{.}%
    {etc}%
    {etc.\@\xspace}%
}
\makeatother
\newcommand{\etal}{\textit{et al}. }

\begin{document}
\pagestyle{headings}
\mainmatter
\def\ECCVSubNumber{2515} 

\title{Temporal Aggregate Representations\\
for Long-Range Video Understanding} 

\titlerunning{Temporal Aggregate Representations for Long-Range Video Understanding}
%
\author{Fadime Sener\inst{1,2} \and
Dipika Singhania\inst{2}\and
Angela Yao\inst{2}}
\authorrunning{Sener et al.}
\institute{University of Bonn, Germany \and
National University of Singapore \\
\email{\{sener@cs.uni-bonn.de\},\{dipika16,ayao\}@comp.nus.edu.sg}}
\maketitle

\begin{abstract}
Future prediction, especially in long-range videos, requires reasoning from current and past observations. In this work, we address questions of temporal extent, scaling, and level of semantic abstraction with a flexible multi-granular temporal aggregation framework.  We show that it is possible to achieve state of the art in both next action and dense anticipation with simple techniques such as max-pooling and attention. To demonstrate the anticipation capabilities of our model, we conduct experiments on Breakfast, 50Salads, and EPIC-Kitchens datasets, where we achieve state-of-the-art results.  With minimal modifications, our model can also be extended for video segmentation and action recognition.
\end{abstract}
 
\section{Introduction}
We tackle long-range video understanding, specifically anticipating not-yet observed but upcoming actions. When developing intelligent systems, one needs not only to recognize what is \emph{currently} taking place -- but also predict what will happen \emph{next}. Anticipating human actions is essential for applications such as smart surveillance, autonomous driving, and assistive robotics. 
 
While action anticipation is a niche (albeit rapidly growing) area, the key issues that arise are germane to long-range video understanding as a whole. How should temporal or sequential relationships be modelled? What temporal extent of information and context needs to be processed? At what temporal scale should they be derived, and how much semantic abstraction is required? The answers to these questions are not only entangled with each other but also depend very much on the videos being analyzed. Here, one needs to distinguish between clipped actions, \eg of UCF101~\cite{Soomro101}, versus the multiple actions in long video streams, \eg of the Breakfast~\cite{kuehne2014language}. In the former, the actions and video clips are on the order of a few seconds, while in the latter, it is several minutes. As such, temporal modelling is usually not necessary for simple action recognition~\cite{huang2018makes}, but more relevant for understanding complex activities~\cite{richard2016temporal,sener2018unsupervised}. 

Temporal models that are built into the architecture~\cite{ding2018weakly,farha2019ms,huang2016connectionist,richard2017weakly} are generally favoured because they allow frameworks to be learned end-to-end. However, this means that the architecture also dictates the temporal extent that can be accounted for. This tends to be short, either due to difficulties in memory retention or model size. As a result, the context for anticipation can only be drawn from a limited extent of recent observations, usually on the order of seconds~\cite{lan2014hierarchical,vondrick2016anticipating,miech2019leveraging}. This, in turn, limits the temporal horizon and granularity of the prediction. 
 
One way to ease the computational burden, especially under longer temporal extents, is to use higher-level but more compact feature abstractions, \eg by using detected objects, people~\cite{lfb2019} or sub-activity labels~\cite{abu2018will,Ke_2019_CVPR} based on the outputs of video segmentation algorithms~\cite{richard2017weakly}. Such an approach places a heavy load on the initial task of segmentation and is often difficult to train end-to-end. Furthermore, since labelling and segmenting actions from video are difficult tasks, their errors may propagate onwards when anticipating future actions. 

Motivated by these questions of temporal modelling, extent, scaling, and level of semantic abstraction, we propose a general framework for encoding long-range video. We aim for flexibility in frame input, \ie ranging from low-level visual features to high-level semantic labels, and the ability to meaningfully integrate recent observations with long-range context in a computationally efficient way. To do so, we split video streams into snippets of equal length and max-pool the frame features within the snippets. We then create ensembles of multi-scale feature representations that are aggregated bottom-up based on scaling and temporal extent. Temporal aggregation~\cite{kline1995computing} is a form of summarization used in database systems. Our framework is loosely analogous as it summarizes the past observations through aggregation, so we name it ``temporal aggregates''.  We summarize our main contributions as follows:
\begin{itemize}
\item We propose a simple and flexible single-stage framework of multi-scale temporal aggregates for videos by relating recent to long-range observations. 
\item Our representations can be applied to several video understanding tasks; in addition to action anticipation, it can be used for recognition and segmentation with minimal modifications and is able to achieve competitive results.
\item Our model has minimal constraints regarding the type of anticipation (dense or next action), type of the dataset (instructional or daily activities), and type of input features (visual features or frame-level labels). 
\item We conduct experiments on Breakfast~\cite{kuehne2014language}, 50Salads~\cite{stein2013combining} and EPIC-Kitchens~\cite{damen2018scaling}. 
\end{itemize}

\section{Related Works}
\textbf{Action recognition} has advanced significantly with deep networks in recent years. Notable works include two steam networks~\cite{simonyan2014two,wang2016temporal}, 3D convolutional networks~\cite{tran2015learning,carreira2017quo}, and RNNs~\cite{donahue2015long,yue2015beyond}. These methods have been designed to encode clips of a few seconds and are typically applied to the classification of \emph{trimmed} videos containing a single action~\cite{Soomro101,kay2017kinetics}. In our paper, we work with long \emph{untrimmed} sequences of complex activities. Such long videos are not simply a composition of independent short actions, as the actions are related to each other with sequence dynamics. Various models for complex activity understanding have been addressed before~\cite{ding2018weakly,farha2019ms,sener2018unsupervised}; these approaches are designed to work on instructional videos by explicitly modelling their sequence dynamics. These models are not flexible enough to be extended to daily activities with loose orderings. Also, when only partial observations are provided, \eg for anticipation, these models cannot be trained in a single stage. 

\textbf{Action anticipation} aims to forecast actions before they occur.   Prior works in immediate anticipation were initially limited to movement primitives like \emph{reaching}~\cite{Koppula15pami} or interactions such as \emph{hugging}~\cite{vondrick2016anticipating}. \cite{mahmud2017joint} presents a model for predicting both the next action and its starting position. \cite{damen2018scaling} presents a large daily activities dataset, along with a challenge for anticipating the next action one second before occurrence. \cite{miech2019leveraging} proposes next action anticipation from recent observations. Recently, \cite{furnari2019rulstm} proposed using an LSTM to summarize the past and another LSTM for future prediction. These works assume near-past information, whereas we make use of long-range past.

\textbf{Dense anticipation} predicts actions multiple steps into the future. Previous methods~\cite{abu2018will,Ke_2019_CVPR} to date require having already segmented temporal observations. Different than these, our model can perform dense anticipation in a single stage without any pre-segmented nor labelled inputs. 

The \textbf{role of motion and temporal dynamics} has been well-explored for video understanding, though the focus has been on short clips~\cite{lin2019tsm,carreira2017quo,huang2018makes}. Some works use longer-term temporal contexts, though still in short videos  ~\cite{li2017temporal,nonlocalNetVLAD}. Recently, Wu~\etal\cite{lfb2019} proposed integrating long-term features with 3D CNNs in short videos and showed the importance of temporal context for action recognition. Our model is similar in spirit to~\cite{lfb2019} in that we couple the recent with the long-range past using attention. One key difference is that we work with ensembles of multiple scalings and granularities, whereas~\cite{lfb2019} work at a single frame-level granularity. As such, we can handle long videos up to tens of minutes, while they are only able to work on short videos. Recently, Feichtenhofer~\etal\cite{feichtenhofer2019slowfast} proposed SlowFast networks, which, similar to our model, encode time-wise multi-scale representations.  These approaches are limited to short videos and cannot be extended to minutes-long videos due to computational constraints. 

\begin{figure} 
\centering 
\includegraphics[width=1\columnwidth]{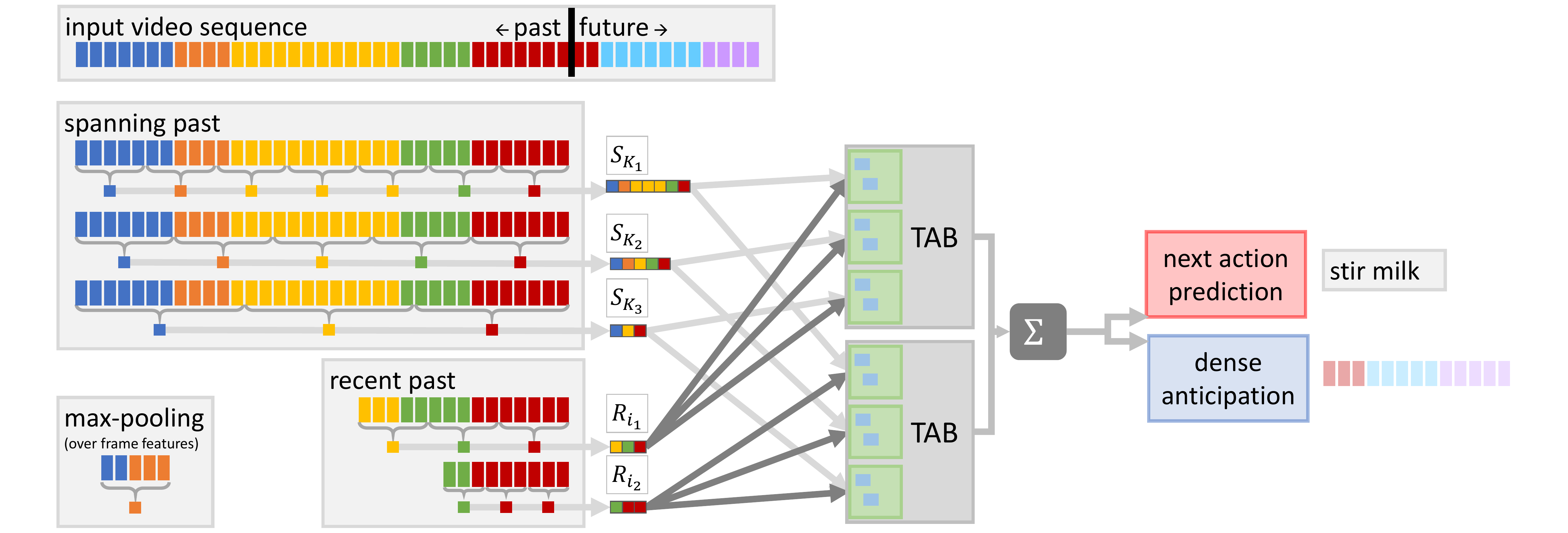}
\caption{Model overview: In this example we use 3 scales for computing the ``spanning past'' features $\mathbf{S}_{K_1}, \mathbf{S}_{K_2}, \mathbf{S}_{K_3}$, and 2 starting points to compute the ``recent past'' features, $\mathbf{R}_{i_1}, \mathbf{R}_{i_2}$, by max-pooling over the frame features in each snippet. Each recent snippet is coupled with all the spanning snippets in our Temporal Aggregation Block (TAB). An ensemble of TAB outputs is used for dense or next action anticipation.}
\label{fig:overview_new}
\end{figure}

\section{Representations}
We begin by introducing the representations, which are inputs to the building blocks of our framework, see Fig.~\ref{fig:overview_new}. We had two rationales when designing our network. First, we relate recent to long-range observations, since some past actions directly determine future actions. Second, to represent recent and long-range past at various granularities, we pool snippets over multiple scales. 
  
\subsection{Pooling} 
For a video of length $T$, we denote the feature representation of a single video frame indexed at time $t$ as $f_t \in \mathbb{R}^{D}, 1\leq t \leq T$. $f_t$ can be derived from low-level features, such as IDT~\cite{wang2013action} or I3D~\cite{carreira2017quo}, or high-level abstractions, such as sub-activity labels derived from segmentation algorithms. To reduce computational load, we work at a snippet-level and define a snippet feature $\textbf{F}_{ij;K}$ as the concatenation of max-pooled features from $K$ snippets, where snippets are partitioned consecutively from frames $i$ to $j$: 
\begin{align}\label{eqn:maxpool}
\begin{split} 
\textbf{F}_{ij;K} & = [F_{i,i+k}, F_{i+k+1,i+2k}, ..., F_{j-k+1,j}], \; \text{where} \\
 ( F_{p,q})_d & = \max\limits_{p \leq t \leq q} \{ f_{t}\}_d\,, \; 1\leq d \leq D,\;\; k=\sfrac{(j-i)}{K}.
\end{split} 
\end{align}
\noindent Here, $F_{p,q}$ indicates the maximum over each dimension $d$ of the frame features in a given snippet between frames $p$ and $q$, though it can be substituted with other alternatives. In the literature, methods representing snippets or segments of frames range from simple sampling and pooling strategies to more complex representations such as learned pooling~\cite{lin2018nextvlad} and LSTMs~\cite{ostyakov2018label}. Especially for long snippets, it is often assumed that a learned representation is necessary~\cite{girdhar2017actionvlad,lee20182nd}, though their effectiveness over simple pooling is still controversial~\cite{wang2016temporal}. The learning of novel temporal pooling approaches goes beyond the scope of this work and is an orthogonal line of development. We verify established methods (see Sec.~\ref{sec:ablations}) and find that a simple max-pooling is surprisingly effective and sufficient. 

\subsection{Recent vs. Spanning Representations}  
Based on different start and end frames $i$ and $j$ and number of snippets $K$, we define two types of snippet features: \emph{``recent''} features $\{\mathcal{R}\}$ from recent observations, and \emph{``spanning''} features $\{\mathcal{S}\}$ drawn from the entire video. The recent snippets cover a couple of seconds (or up to a minute, depending on the temporal granularity) before the current time point while spanning snippets refer to the long-range past and may last up to ten minutes. For \emph{``recent''} snippets, the end frame $j$ is fixed to the current time point $t$, and the number of snippets is fixed to $K_R$. The recent snippet features $\mathcal{R}$ can be defined as a feature bank of snippet features with different start frames $i$, \ie
\begin{align}\label{eqn:recentsnippet}
\begin{split}
\mathcal{R} & = \{\mathbf{F}_{i_{1}t;K_R}, \mathbf{F}_{i_{2}t;K_R}, ..., \mathbf{F}_{i_Rt;K_R}\} = \{\mathbf{R}_{i_1}, \mathbf{R}_{i_2}, ..., 
\mathbf{R}_{i_R}\},
\end{split}
\end{align}
where $\mathbf{R}_i \in \mathbb{R}^{D \times K_R} $ is a shorthand to denote $\mathbf{F}_{i\,t;K_R}$, since endpoint $t$ and number of snippets $K_R$ are fixed. In Fig.~\ref{fig:overview_new} we use two starting points to compute the ``recent'' features and represent each with $K_R\!=\!3$ snippets (\tikz{\snippetBox{boxYellow}{0}\snippetBox{boxGreen}{1.25}\snippetBox{boxRed}{2.5}} \& \tikz{\snippetBox{boxGreen}{0}\snippetBox{boxRed}{1.25}\snippetBox{boxRed}{2.5}}).

For \emph{``spanning''} snippets, $i$ and $j$ are fixed to the start of the video and current time point,~\ie $i\!=\!0, j\!=\!t$. Spanning snippet features $\mathcal{S}$ are defined as a feature bank of snippet features with varying number of snippets $K$, \ie 
\begin{align}\label{eqn:spanningsnippet}
\begin{split}
\mathcal{S} & = \{\mathbf{F}_{0\,t;K_1}, \mathbf{F}_{0\,t;K_2}, ..., \mathbf{F}_{0\,t;K_S}\} = \{\mathbf{S}_{K_1}, \mathbf{S}_{K_2}, ..., 
\mathbf{S}_{K_S}\},
\end{split}
\end{align}
\noindent where $\mathbf{S}_K \in \mathbb{R}^{D \times K} $ is a shorthand for $\mathbf{F}_{0\,t;K}$. In Fig.~\ref{fig:overview_new} we use three scales to compute the ``spanning'' features with $K = \{7,5,3\}$ (\tikz{\snippetBox{boxBlue}{0}\snippetBox{boxOrange}{1.25}\snippetBox{boxYellow}{2.5}\snippetBox{boxYellow}{3.75}\snippetBox{boxYellow}{5.0}\snippetBox{boxGreen}{6.25}\snippetBox{boxRed}{7.5}},\tikz{\snippetBox{boxBlue}{0}\snippetBox{boxOrange}{1.25}\snippetBox{boxYellow}{2.5}\snippetBox{boxGreen}{3.75}\snippetBox{boxRed}{5.0}} \&\tikz{\snippetBox{boxBlue}{0}\snippetBox{boxYellow}{1.25}\snippetBox{boxRed}{2.5}}).

Key to both types of representations is the ensemble of snippet features from multiple scales. We achieve this by varying the number of snippets $K$ for the spanning past. For the recent past, it is sufficient to keep the number of snippets $K_R$ fixed, and vary only the start point $i$, due to redundancy between $\mathcal{R}$ and $\mathcal{S}$ for the snippets that overlap. For our experiments, we work with snippets ranging from seconds to several minutes.

\begin{figure*}[!t] 
\centering 
\includegraphics[width=1\textwidth]{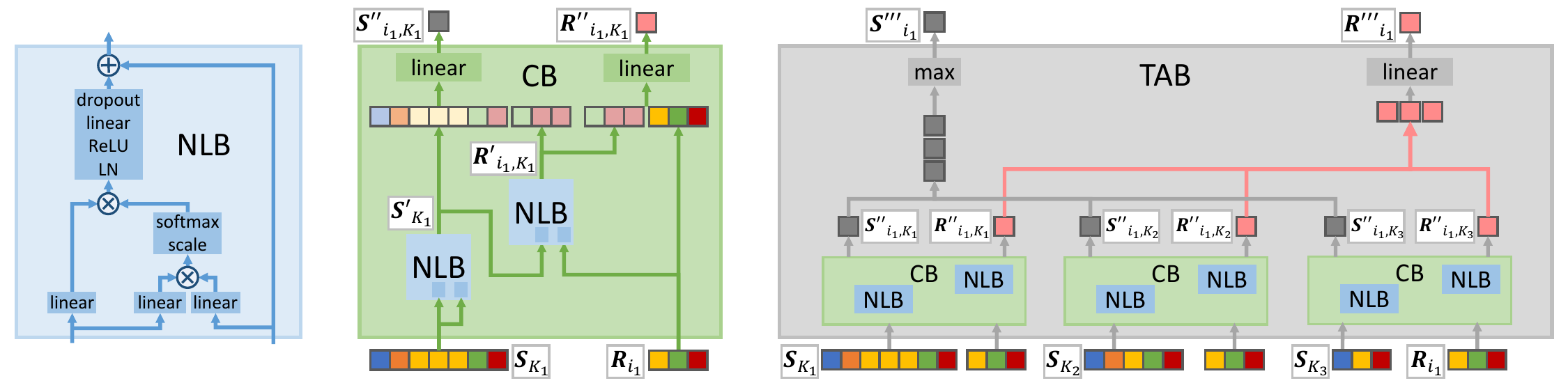} 
\caption{Model components: Non-Local Blocks (NLB) compute interactions between two representations via attention (Sec.~\ref{sec:nlb}). Two such NLBs are combined in a Coupling Block (CB), which calculates the attention-reweighted spanning and recent representations (Sec.~\ref{sec:cb}). We couple each recent with all spanning representations via individual CBs and combine their outputs in a Temporal Aggregation Block (TAB) (Sec.~\ref{sec:tab}). The outputs of multiple such TABs are combined to perform anticipation,  Fig.~\ref{fig:overview_new}.}
\label{fig:overview} 
\end{figure*} 

\section{Framework}
In Fig.~\ref{fig:overview} we present an overview of the components used in our framework, which we build in a bottom-up manner, starting with the recent and spanning features $\mathcal{R}$ and $\mathcal{S}$, which are coupled with non-local blocks (NLB) (Sec.~\ref{sec:nlb}) within coupling blocks (CB) (Sec.~\ref{sec:cb}). The outputs of the CBs from different scales are then aggregated inside temporal aggregation blocks (TAB) (Sec.~\ref{sec:tab}). Outputs of different TABs can then be chained together for either next action anticipation or dense anticipation (Secs.~\ref{sec:instructional_act},~\ref{sec:daily_act}).

\subsection{Non-Local Blocks (NLB)}\label{sec:nlb} 
We apply non-local operations to capture relationships amongst the spanning snippets and between spanning and recent snippets. Non-local blocks~\cite{wang2018non} are a flexible way to relate features independently from their temporal distance and thus capture long-range dependencies. We use the modified non-local block from~\cite{lfb2019}, which adds layer normalization~\cite{ba2016layer} and dropout~\cite{srivastava2014dropout} to the original one in~\cite{wang2016temporal}.   Fig.~\ref{fig:overview} (left) visualizes the architecture of the block, the operation of which we denote as $\text{NLB}(\cdot,\cdot)$.

\subsection{Coupling Block (CB)}\label{sec:cb} 
Based on the NLB, we define attention-reweighted spanning and recent outputs:
\begin{align}
 \textbf{S}'_{K} = NLB(\textbf{S}_{K},\textbf{S}_{K})
\quad \text{and} \quad 
 \textbf{R}'_{i,K} = NLB(\textbf{S}'_{K}, \textbf{R}_i)\label{coupledAttn}.
\end{align}
\noindent $\textbf{R}'_{i,K}$ is coupled with either $\textbf{R}_i$ or $\textbf{S}'_K$ via concatenation and a linear layer. This results in the fixed-length representations $\textbf{R}''_{i,K}$ and $\textbf{S}''_{i,K}$, where $i$ is the starting point of the recent snippet and $K$ is the scale of the spanning snippet. 
 
\subsection{Temporal Aggregation Block (TAB)}\label{sec:tab}
The final representation for recent and spanning past is computed by aggregating outputs from multiple CBs. For the same recent starting point $i$, we concatenate $\textbf{R}''_{i,K_1}, ...,$ $\textbf{R}''_{i,K_S}$ for all spanning scales and pass the concatenation through a linear layer to compute $\textbf{R}'''_i$. The final spanning representation $\textbf{S}'''_{i}$ is a max over all $\textbf{S}''_{i,K_1}, ..., \textbf{S}''_{i,K_S}$. We empirically find that taking the max outperforms other alternatives like linear layers and/or concatenation for the spanning past (Sec. \ref{sec:ablations}).  TAB outputs, by varying recent starting points $\{i\}$ and scales of spanning snippets $\{K\}$, are multi-granular video representations that aggregate and encode both the recent and long-range past. We name these \textbf{temporal aggregate representations}. Fig.\ref{fig:overview_new} shows an example with 2 recent starting points and 3 spanning scales. These representations are generic and can be applied in various video understanding tasks (see Sec.~\ref{sec:pred}) from long streams of video. 
 
\subsection{Prediction Model}\label{sec:pred}
\begin{figure}[t]
\centering 
\includegraphics[width=0.89\columnwidth]{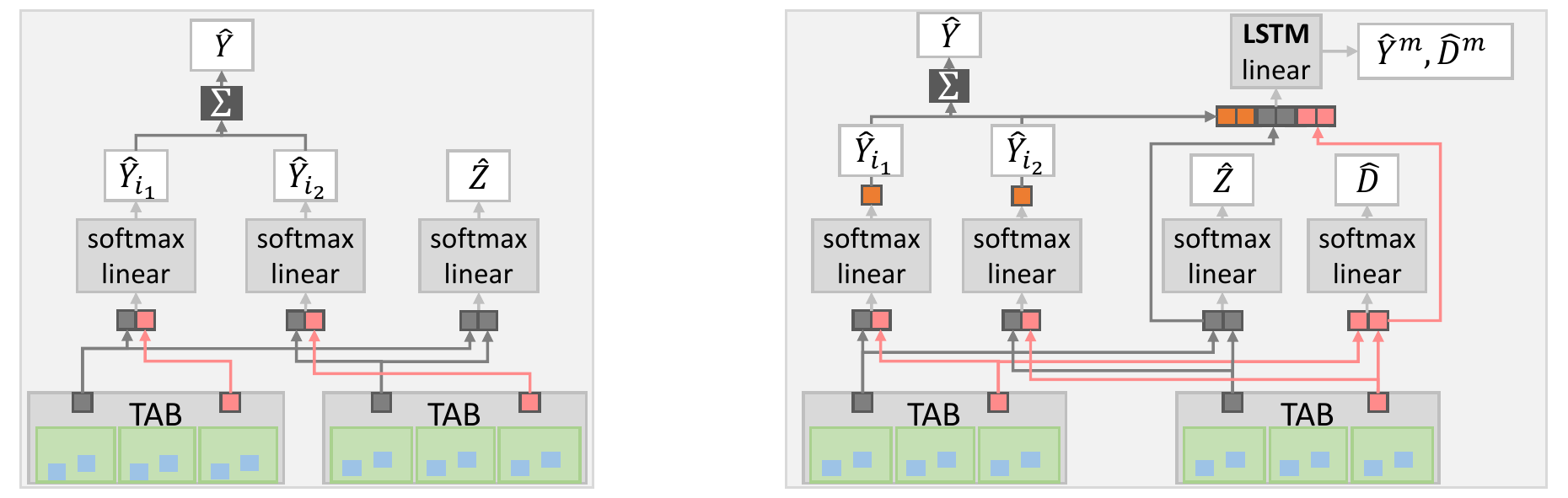}
\caption{Prediction models for classification (left) and sequence prediction (right).} 
\label{fig:next_action_pred} 
\end{figure}

\textbf{Classification:} For single-label classification tasks such as next action anticipation, temporal aggregate representations can be used directly with a classification layer (linear + softmax). A cross-entropy loss based on ground truth labels $Y$ can be applied to the predictions $\hat{Y}_{i}$, where $Y$ is either the current action label for recognition, or the next action label for next action prediction (see Fig.~\ref{fig:next_action_pred}). 

When the individual actions compose a complex activity (\eg ``take bowl'' and ``pour milk'' as part of ``making cereal''  in Breakfast~\cite{kuehne2014language}), we can add an additional loss based on the complex activity label $Z$. Predicting $Z$ as an auxiliary task helps with anticipation. For this we concatenate $\textbf{S}'''_{i_1}, \ldots, \textbf{S}'''_{i_R}$ from all TABs and pass them through a classification layer to obtain $\hat{Z}$. The overall loss is a sum of the cross entropies over the action and complex activity labels: 
\begin{align}\label{eq:loss_next_action}
 \mathcal{L}_{\text{cl}} = \mathcal{L}_{\text{comp}} + \mathcal{L}_{\text{action}}
 = - \sum_{n=1}^{N_Z} Z_n \log ( \hat{Z} )_n - \sum_{r=1}^{R} \sum_{n=1}^{N_Y} Y_n \log ( \hat{Y}_{i_r} )_n,
\end{align}
where $i_r$ is one of the R recent starting points, and $N_Y$ and $N_Z$ are the total number of actions and complex activity classes respectively. During inference, the predicted scores   are summed for a final prediction, \ie $\hat{Y}\!=\! \max_n(\sum_{r = 1}^R \hat{Y}_{i_r})_n\label{inference_next}$. 

We frame sequence segmentation as a classification task and predict frame-level action labels of complex activities. Multiple sliding windows with fixed start and end times are generated and then classified into actions using Eq.\ref{eq:loss_next_action}. 
 
\textbf{Sequence prediction:} The dense anticipation task predicts frame-wise actions of the entire future sequence. Previously,~\cite{abu2018will} predicted future segment labels via classification and regressed the durations. We opt to estimate both via classification. The sequence duration is discretized into $N_D$ intervals and represented as one-hot encodings $D\in \{0,1\}^{N_D}$.  For dense predictions, we perform multi-step estimates.  We first estimate the current action and complex activity label, as per Eq.~\ref{loss_next_action}. The current duration $D$ is then estimated via a classification layer applied to the concatenation of recent temporal aggregates $\textbf{R}'''_{i_1}, ..., \textbf{R}'''_{i_R}$. 

For future actions, we concatenate all recent and spanning temporal aggregates $\textbf{R}'''_{i_1}, ..., \textbf{R}'''_{i_R}$ and $\textbf{S}'''_{i_1}, ..., \textbf{S}'''_{i_R}$ and the classification layer outputs $\hat{Y}_{i_1}, ..., \hat{Y}_{i_R}$, and pass the concatenation through a linear layer before feeding the output to a one-layer LSTM. The LSTM predicts at each step $m$ an action vector $\hat{Y}^{m}$ and a duration vector $\hat{D}^{m}$ (see Fig.~\ref{fig:next_action_pred}).  The dense anticipation loss is a sum of the cross-entropies over the current action, its duration, future actions and durations, and task labels respectively: 
{
{
\begin{align}\label{loss_next_action}
\mathcal{L}_{\text{dense}}\!=\! 
\mathcal{L}_{\text{cl}}\!-\!\sum_{n=1}^{N_D} D_n \log ( \hat{D} )_n\!
-\!\frac{1}{M}\sum_{m=1}^{M}\left(
\sum_{n=1}^{N_Y} Y^{m}_{n}\log(\hat{Y}^{m})_{n} \!+\!
\sum_{n=1}^{N_D} D^{m}_{n}\log(\hat{D}^{m})_{n}\right)
\end{align}
}
}
During inference we sum the predicted scores (post soft-max) for all starting points $i_r$ to predict the current action as $\max_n(\sum_{r = 1}^R \hat{Y}_{i_r})_n$.
The LSTM is then applied recurrently to predict subsequent actions and durations.  
 
\section{Experiments}
\subsection{Datasets and Features}\label{sec:datasets}
We experiment on  Breakfast~\cite{kuehne2014language}, 50Salads~\cite{stein2013combining} and EPIC-Kitchens~\cite{damen2018scaling}. The sequences in each dataset reflect realistic durations and orderings of actions, which is crucial for real-world deployment of anticipation models. Relevant datasets statistics  are given in Table~\ref{tab:my_label}. One notable difference between these datasets is the label granularity; it is very fine-grained for EPIC, hence their 2513 action classes, versus the coarser 48 and 17 actions of Breakfast and 50Salads. As a result, the median action segment duration is 8-16x shorter.
 
Feature-wise, we use pre-computed Fisher vectors~\cite{abu2018will} and I3D~\cite{carreira2017quo} for Breakfast, Fisher vectors for 50Salads, and appearance, optical flow and object-based features  provided by~\cite{furnari2019rulstm} for EPIC. Results for Breakfast and 50Salads are averaged over the predefined 4 and 5 splits respectively. Since 50Salads has only a single complex activity (making salad) we omit complex activity prediction for it. For EPIC, we report results on the test set. Evaluation measures are class accuracy (Acc.) for next action prediction and mean over classes~\cite{abu2018will} for dense prediction. We report Top-1 and Top-5 accuracies to be consistent with~\cite{miech2019leveraging,furnari2019rulstm}. 

Hyper-parameters for spanning $\{K\}$, recent scales $K_R$ and recent starting points $\{i\}$ are given in Table~\ref{tab:my_label}. We cross validated the parameters on different splits of 50Salads and Breakfast; on EPIC, we select parameters with the validation set~\cite{furnari2019rulstm}. 

\begin{table}[t!]
\centering
\resizebox{\columnwidth}{!}{ 
\setlength{\tabcolsep}{2pt}
\begin{tabular}{|c|c|c|c|c|c|c|}
\hline
Dataset& \begin{tabular}[c]{@{}c@{}}video duration\\ median, mean $\pm$std\end{tabular} & \# classes& \# segments & $\{i\}$(in seconds) & $K_R$ & $\{K\}$\\
\hline
Breakfast(@15fps) & 15.1s, 26.6s $\pm$36.8 & 48 & 11.3K & $\{t-10,t-20,t-30\}$ & 5 & $\{10,15,20\}$\\
50Salads(@30fps) & 29.7s, 38.4s $\pm$31.5 & 17 & 0.9K & $\{t-5,t-10,t-15\}$ & 5 & $\{5,10,15\}$\\
EPIC(@60fps) & 1.9s, 3.7s $\pm$ 5.6 & 2513 & 36.6K& $\{t\!-\!1.6,t\!-\!1.2,t\!-\!0.8,t\!-\!0.4\}$ & 2 & $\{2,3,5\}$\\
\hline
\end{tabular}}
\caption{Dataset details and our respective model parameters.}
\label{tab:my_label}
\end{table} 

\subsection{Component validation}\label{sec:ablations} 
We verify each component's utility via a series of ablation studies summarized in Table~\ref{tab:componentAblations}. As our main motivation was to develop a representation for anticipation in long video streams, we validate on Breakfast for next action anticipation.  Our full model gets a performance of 40.1\% accuracy averaged over actions. 

\begin{table}[t]
\centering
\resizebox{\columnwidth}{!}{ 
\setlength{\tabcolsep}{4pt}
\begin{tabular}{|l|l|l|l|l|l|l|}
\hline
    Pooling type & \cellcolor{magenta!15} frame sampling & \cellcolor{magenta!15}GRU & \cellcolor{magenta!15}BiLSTM &  \multicolumn{2}{l|}{\cellcolor{magenta!15}mean-pooling} & \cellcolor{magenta!15}max-pooling \\ \cline{1-1}
    Acc. & \cellcolor{magenta!15}32.1 & \cellcolor{magenta!15}37.9 & \cellcolor{magenta!15}38.7 &  \multicolumn{2}{l|}{\cellcolor{magenta!15}36.6} & \cellcolor{magenta!15}\textbf{40.1} \\ \hline\hline
    Influence of & \multicolumn{5}{l|}{Changes in components} & Acc. (Drop) \\ \hline
    Non-Local Blocks (NLB) & \multicolumn{5}{l|}{\cellcolor{blue!10}replace all NLBs with concatenation + linear layer} & \cellcolor{blue!10}33.7 (6.4\% ) \\ \hline
    \multirow{3}{*}{Coupling Blocks (CB)} 
    & \multicolumn{5}{l|}{\cellcolor{green!10}only couple the $\textbf{S}_{K}$ and $\textbf{S}_{K}$ in CBs} & \cellcolor{green!10}35.1 (5.0\% )\\ \cline{2-7} 
    & \multicolumn{5}{l|}{\cellcolor{green!10}only couple the $\textbf{R}_i$ and $\textbf{R}_i$ in CBs } & \cellcolor{green!10}34.2 (5.9\% )\\ \cline{2-7} 
    & \multicolumn{5}{l|}{\cellcolor{green!10}replace CBs with concatenation + linear layer} & \cellcolor{green!10}33.4 (6.7\% )\\ \hline
    \multirow{3}{*}{\begin{tabular}[c]{@{}l@{}}Temporal Aggregation\\ Blocks (TAB)\end{tabular}} 
    & \multicolumn{5}{l|}{\cellcolor{gray!20}a single CB is used in TABs} & \cellcolor{gray!20}38.0 (2.1\% ) \\ \cline{2-7} 
    & \multicolumn{5}{l|}{\cellcolor{gray!20}three CBs are used in a single TAB} & \cellcolor{gray!20}37.7 (2.4\% ) \\ \cline{2-7} 
    & \multicolumn{5}{l|}{\cellcolor{gray!20}a single a CB is used without any TABs} & \cellcolor{gray!20}32.1 (8.0\% ) \\ \hline
\end{tabular}}\\
\resizebox{\columnwidth}{!}{ 
\setlength{\tabcolsep}{0.1pt}
\begin{tabular}{|cll|ccccccc|} 
\hline
\multirow{6}{*}{\begin{tabular}[c]{@{}c@{}}Recent \&\\ Spanning \\ Repr.\end{tabular}} & \multicolumn{1}{c}{\multirow{2}{*}{\textbf{(a)}}} & starting points $i$ & $i_1=t-10$ & $i_2=t-20$ & $i_3=t-30$ & $i_4=0$ & \textbf{$\mathbf{\{i_1, i_2,i_3\}}$} &&\\ 
& \multicolumn{1}{c}{} & Acc. & 36.9& 37.7& 37.2& 35.1 & 40.1 &&\\ \cline{2-10}
& \multirow{2}{*}{\textbf{(b)}} & spanning scales $K$ &$\{5\}$ & $\{10\}$& $\{15\}$& $\{20\}$ & $\{10,15\}$ & $\{$10,15,20$\}$ & $\{$5,10,15,20$\}$ \\ 
&& Acc. & 37.4& 38.0& 37.5& 37.4 & 39.0 & \textbf{40.1} & 40.2 \\ \cline{2-10} 
& \multirow{2}{*}{\textbf{(c)}} & recent scales $K_R$ &$1$ & $3$ & \textbf{5} & $10$ &&&\\ 
&& Acc. & 38.7& 39.5& \textbf{40.1}& 38.6 &&&\\\hline
\end{tabular}}
\caption{Ablations on the influence of different model components.}
\label{tab:componentAblations} 
\end{table}

\textbf{Video Representation:} Several short-term feature representations have been proposed for video, \eg~3D convolutions~\cite{tran2015learning}, or combining CNNs and RNNs for sequences~\cite{yue2015beyond,donahue2015long}. For long video streams, it becomes difficult to work with all the raw features. Selecting representative features can be as simple as sub-sampling the frames~\cite{feichtenhofer2019slowfast,xiao2020audiovisual}, or pooling~\cite{wang2016temporal}, to more complex RNNs~\cite{yue2015beyond}. Current findings in the literature are not in agreement. Some propose learned strategies~\cite{miech2017learnable,lee20182nd}, while others advocate pooling~\cite{wang2016temporal}. Our experiments align with the latter, showing that max-pooling is superior to both sampling (+8\%) and the GRU (+2.2\%) or bi-directional LSTM~\cite{conneau2017supervised} (+1.4\%). The performance of GRU and BiLSTM are comparable to average-pooling, but require much longer training and inference time. For us, max-pooling works better than average pooling; this contradicts the findings of~\cite{wang2016temporal}. We attribute this to the fact that we pool over minutes-long snippets and it is likely that mean- smooths away salient features that are otherwise preserved by max-pooling. We conducted a similar ablations on EPIC, where we observed a 1.3\% increase with max- over mean-pooling.

\textbf{Recent and Spanning Representations:}
In our ablations, unless otherwise stated, an ensemble of 3 spanning scales $K\!=\!\{10,15,20\}$ and 3 recent starting points $i\!=\!\{t\!-\!10,t\!-\!20,t\!-\!30\}$ are used.  Table~\ref{tab:componentAblations} \textbf{(a)} compares single starting points for the recent snippet features versus an ensemble. With a single starting point, points too near to and too far from the current time decrease the performance. The worst individual result is with $i_4 = 0$, \ie using the entire sequence; the peak is at $i_2 = t - 20$, though an ensemble is still best. In Table~\ref{tab:componentAblations} \textbf{(b)}, we show the influence of spanning snippet scales. These scales determine the temporal snippet granularity; individually, results are not significantly different across the scales, but as we begin to aggregate an ensemble, the results improve. The ensemble with 4 scales is best but only marginally better than 3, at the expense of a larger network, so we choose $K\!=\!\{10,15,20\}$. In Table~\ref{tab:componentAblations} \textbf{(c)}, we show the influence of recent snippet scales, we find $K_R = 5$ performs best.
 
\textbf{Coupling Blocks:} Previous studies on simple video understanding have shown the benefits of using features from both the recent and long-range past~\cite{li2017temporal,lfb2019}. A na\"ive way to use both is to simply concatenate, though combining the two in a learned way, \eg via attention, is superior (+6.4\%). To incorporate attention, we apply NLBs~\cite{wang2018non}, which is an adaptation of the attention mechanism that is popularly used in machine translation. When we replace our CBs with concatenation and a linear layer, there is a drop of 6.7\%. When we do not use coupling but separately pass the $\textbf{R}_i$ and $\textbf{S}_{K}$ through concatenation and a linear layer, there is a drop of 7.5\%. We find also that coupling the recent $\textbf{R}_i$ and long range $\textbf{S}_{K}$ information is critical. Coupling only recent information (-5.9\%) does not keep sufficient context, whereas coupling only long-range past (-5\%) does not leave sufficient representation for the more relevant recent aspects. 

\textbf{Temporal Aggregation Blocks} (TAB) are the most critical component. Omitting them and classifying a single CB's outputs significantly decreases accuracy (-8\%). The strength of the TAB comes from using ensembles of coupling blocks as input (single, -2.1\%) and using the TABs in an ensemble (single, -2.4\%). 
 
\textbf{Additional ablations:} When we omit the auxiliary complex activity prediction, \ie removing the $Z$ term from Eq.~\ref{loss_next_action} (``no Z''), we observe a slight performance drop of 1.1\%. In our model we max pool over all $\textbf{S}''_{i,K_1}, ..., \textbf{S}''_{i,K_S}$ in our TABs. When we replace the max-pooling with concatenation + linear, we reach an accuracy of 37.4.
We also try to disentangle the ensemble effect from the use of multi-granular representations. When we fix the spanning past scales $K$ to $\{15,15,15\}$ and all the starting points to $i=t-20$, we observe a drop of 1.2\% in accuracy which indicates the importance of our multi-scale representation.

\begin{table}[t] 
\centering
\resizebox{\columnwidth}{!}{ 
\setlength{\tabcolsep}{15pt}
\begin{tabular}{|l|l|l|l|c|c|c|}
\hline
Method & Input & Segmentation Method and Feature & Breakfast & 50Salads \\\hline
\cite{vondrick2016anticipating} & FC7 features & - & 8.1 & 6.2\\
\cite{miech2019leveraging} & R(2+1)D & - & 32.3 & \\\hline 
\rowcolor{red!15}RNN \cite{abu2018will} & segmentation & ~\cite{richard2017weakly}, Fisher & \textbf{30.1} & 30.1 \\
\rowcolor{red!15}CNN \cite{abu2018will} & segmentation & ~\cite{richard2017weakly}, Fisher & 27.0 & 29.8 \\
\rowcolor{red!15}ours no $Z$ & Fisher & - & 29.2 & \textbf{31.6}\\
\rowcolor{red!15}ours & Fisher & - & 29.7 & \\\hline 
\rowcolor{blue!15}ours & I3D & - & 40.1 & 40.7 \\ 
\rowcolor{blue!15}ours & segmentation & ours, I3D & 43.1 & \\
\rowcolor{blue!15}ours & segmentation + I3D & ours, I3D & 47.0 & \\\hline 
\rowcolor{green!15}ours & frame GT & - & 64.7 & 63.8\\
\rowcolor{green!15}ours & frame GT + I3D & - & 63.1 & \\ \hline 
\end{tabular} }
\caption{Next action anticipation comparisons on Breakfast and 50Salads, given different frame inputs frame inputs, GT action labels, Fisher vectors and I3D features.}
\label{tab:SOA_next_50Salads} 
\end{table} 

\subsection{Anticipation on Procedural Activities - Breakfast \& 50 Salads} \label{sec:instructional_act}

\subsubsection{Next Action Anticipation}\label{sec:next_action}
predicts the action that occurs 1 second from the current time $t$. We compare to the state of the art in Table \ref{tab:SOA_next_50Salads} with two types of frame inputs: spatio-temporal features (Fisher vectors or I3D) and frame-wise action labels (either from ground truth or via a separate segmentation algorithm) on Breakfast. Compared to previous methods using only visual features as input, we outperform CNN (FC7) features~\cite{vondrick2016anticipating} and spatio-temporal features R(2+1)D~\cite{miech2019leveraging} by a large margin (+32.3\% and +8.1\%). While the inputs are different, R(2+1)D features were shown to be comparable to I3D features~\cite{tran2018closer}. Since~\cite{miech2019leveraging} uses only recent observations, we conclude that incorporating the spanning past into the prediction model is essential.

Our method degrades when we replace I3D with the weaker Fisher vectors (40.1\% vs 29.7\%). Nevertheless, this result is competitive with methods using action labels~\cite{abu2018will} (30.1\% with RNN) derived from segmentation algorithms~\cite{richard2017weakly} using Fisher vectors as input. For fair comparison, we report a variant without the complex activity prediction (``no $Z$''), which has a slight performance drop (-0.5\%).  If we use action labels as inputs instead of visual features, our performance improves from 40.1\% to 43.1\%; merging labels and visual features gives another 4\% boost to 47\%. In this experiment we use segmentation results from our own framework,  
(see Sec.~\ref{sec:video_seg}). However, if we substitute ground truth instead of segmentation labels, there is still a 17\% gap. This suggests that the quality of the segmentation matters. When the segmentation is very accurate, adding additional features does not help and actually slightly deteriorates results (see Table~\ref{tab:SOA_next_50Salads} ``frame GT'' vs. ``frame GT + I3D'').

In Table \ref{tab:SOA_next_50Salads}, we also report results for 50Salads. Using Fisher vectors we both outperform the state of the art by 1.8\% and the baseline with CNN features~\cite{vondrick2016anticipating} by 25.4\%. Using I3D features improves the accuracy by 9.1\% over Fisher vectors.

\begin{table}[t] 
\centering
\resizebox{\columnwidth}{!}{
\setlength{\tabcolsep}{5pt} 
\begin{tabular}{|l|llllllll||llllllll|}
\hline
 & \multicolumn{8}{c||}{Breakfast} & \multicolumn{8}{c|}{50salads} \\ \hline
Obs. & \multicolumn{4}{c|}{20\%} & \multicolumn{4}{c||}{30\%} & \multicolumn{4}{c|}{20\%} & \multicolumn{4}{c|}{30\%} \\ \hline
Pred.& 10\% & 20\% & 30\% & \multicolumn{1}{l|}{50\%}& 10\% & 20\% & 30\% & 50\% & 10\% & 20\% & 30\% & \multicolumn{1}{l|}{50\%}& 10\% & 20\% & 30\% & 50\% \\ \hline
\textbf{A} & \multicolumn{8}{l||}{\cellcolor{green!15} \textbf{Labels} (GT)} & \multicolumn{8}{l|}{\cellcolor{green!15} \textbf{Labels} (GT)}\\ \hline
    \rowcolor{green!15}RNN\cite{abu2018will} & 60.4 & 50.4 & 45.3 & 40.4 & 61.5 & 50.3 & 45.0 & 41.8 & 42.3 & 31.2 & 25.2 & 16.8 & 44.2 & 29.5 & 20.0 & 10.4 \\
    \rowcolor{green!15}CNN\cite{abu2018will} & 58.0 & 49.1 & 44.0 & 39.3 & 60.3 & 50.1 & 45.2 & 40.5 & 36.1 & 27.6 & 21.4 & 15.5 & 37.4 & 24.8 & 20.8 & 14.1 \\
    \rowcolor{green!15}Ke\cite{Ke_2019_CVPR} & 64.5 & \textbf{56.3} & \textbf{50.2} & \textbf{44.0} & 66.0 & 55.9 & \textbf{49.1} & \textbf{44.2} & 45.1 & 33.2 & 27.6 & 17.3 & \textbf{46.4} & \textbf{34.8} & \textbf{25.2} & 13.8 \\
    \rowcolor{green!15}ours & \textbf{65.5} & 55.5 & 46.8 & 40.1 & \textbf{67.4} & \textbf{56.1} & 47.4 & 41.5 &  \textbf{47.2}	& \textbf{34.6}	 & \textbf{30.5}	& \textbf{19.1} & 44.8	& 32.7	 & 23.5	& \textbf{15.3}  \\ \hline
\textbf{B}& \multicolumn{8}{l||}{\cellcolor{red!15} \textbf{Features} (Fisher) }& \multicolumn{8}{l|}{\cellcolor{red!15}  \textbf{Features} (Fisher)} \\ \hline
    \rowcolor{red!15}CNN\cite{abu2018will}& 12.8 & 11.6 & 11.2 & 10.3 & 17.7 & 16.9 & 15.5 & 14.1 &&&&&&&&\\ 
    \rowcolor{red!15} ours& \textbf{15.6} & \textbf{13.1}  & \textbf{12.1}  & \textbf{11.1}  & \textbf{19.5}  & \textbf{17.0}  & \textbf{15.6} & \textbf{15.1} &  {25.5}	 & {19.9} & {18.2}	 & {15.1} & {30.6} &{22.5}	  & {19.1}	 & {11.2}   \\ \hline
\textbf{C}& \multicolumn{8}{l||}{\cellcolor{red!15}   \textbf{Labels} (Fisher + \cite{richard2017weakly} (Acc. 36.8/42.9)) }& \multicolumn{8}{l|}{\cellcolor{red!15}   \textbf{Labels} (Fisher + \cite{richard2017weakly} (Acc.  66.8/66.7)) }\\ \hline
    \rowcolor{red!15}RNN\cite{abu2018will} & 18.1 & \textbf{17.2} & 15.9 & 15.8 & 21.6 & 20.0 & 19.7 & 19.2 & 30.1 & 25.4 & 18.7 & 13.5 & 30.8 & 17.2 & 14.8 & 9.8\\
    \rowcolor{red!15}CNN\cite{abu2018will} & 17.9 & 16.4 & 15.4 & 14.5 & 22.4 & 20.1 & 19.7 & 18.8 & 21.2 & 19.0 & 16.0 & 9.9& 29.1 & 20.1 & 17.5 & 10.9 \\
    \rowcolor{red!15}Ke\cite{Ke_2019_CVPR} & 18.4 & \textbf{17.2} & 16.4 & \textbf{15.8} & 22.8 & \textbf{20.4} & 19.6 & \textbf{19.8} & 32.5 & \textbf{27.6} & 21.3 & \textbf{16.0} & \textbf{35.1} & \textbf{27.1} & 22.1 & 15.6 \\
    \rowcolor{red!15}ours & \textbf{18.8} & 16.9 & \textbf{16.5}& 15.4& \textbf{23.0}& 20.0& \textbf{19.9}& 18.6 & \textbf{32.7} &  26.3 &  \textbf{21.9} &  15.6 &  32.3 & 25.5 &  \textbf{22.7} &  \textbf{17.1} \\ \hline
& \multicolumn{8}{l||}{\cellcolor{red!15} Concatenate B and C}& \multicolumn{8}{l|}{\cellcolor{red!15} Concatenate B and C}\\ \hline
    \rowcolor{red!15} ours& \textbf{25.0} & \textbf{21.9} & \textbf{20.5} & \textbf{18.1} & \textbf{23.0} & \textbf{20.5} & 19.8 &\textbf{19.8} & \textbf{34.7} & 25.9  & \textbf{23.7} &  15.7 & 34.5 & 26.1	& 19.0 &  15.5 \\ \hline
\textbf{D}& \multicolumn{8}{l||}{\cellcolor{blue!15}  \textbf{Features} (I3D)} &&&&&&&&\\
    \cellcolor{blue!15} ours &\cellcolor{blue!15} 24.2 &\cellcolor{blue!15} 21.1 &\cellcolor{blue!15} 20.0	&\cellcolor{blue!15} 18.1	&\cellcolor{blue!15} 30.4	&\cellcolor{blue!15} 26.3	&\cellcolor{blue!15} 23.8	&\cellcolor{blue!15} 21.2 &&&&&&&& \\ \cline{1-9} 
\textbf{E}& \multicolumn{8}{l||}{\cellcolor{blue!15}  \textbf{Labels} (I3D + our seg. (Acc.  54.7/57.8)) } &&&&&&&&\\ 
    \cellcolor{blue!15} ours &\cellcolor{blue!15}    \textbf{37.4}	 &\cellcolor{blue!15}  31.2 &\cellcolor{blue!15}  30.0 &\cellcolor{blue!15} 26.1 &\cellcolor{blue!15} 39.5 &\cellcolor{blue!15}  34.1	 &\cellcolor{blue!15} 31.0	 &\cellcolor{blue!15} \textbf{27.9}    &&&&&&&&\\   \cline{1-9}
    & \multicolumn{8}{l||}{\cellcolor{blue!15} Concatenate D and E} &&&&&&&&\\
    \cellcolor{blue!15} ours&\cellcolor{blue!15} 37.1	&\cellcolor{blue!15} \textbf{31.8}	 &\cellcolor{blue!15} \textbf{30.1}	 &\cellcolor{blue!15} \textbf{27.1}	&\cellcolor{blue!15} \textbf{39.8}	&\cellcolor{blue!15} \textbf{34.2}	 &\cellcolor{blue!15} \textbf{31.9}	&\cellcolor{blue!15} 27.8 &&&&&&&&\\\hline
\end{tabular} }  
\caption{Dense anticipation mean over classes on Breakfast and 50salads, given different frame inputs frame inputs, GT action labels, Fisher vectors and I3D features.}
\label{tab:SOA_dense_breakfast_50salad}
\end{table} 
 
\subsubsection{Dense Anticipation}\label{sec:dense_ant}
predicts frame-wise actions; accuracies are given for specific portions of the remaining video (Pred.) after observing a given percentage of the past (Obs.). We refer the reader to the supplementary for visual results. Competing methods \cite{abu2018will} and \cite{Ke_2019_CVPR} have two stages; they first apply temporal video segmentation and then use outputs~\cite{richard2017weakly}, \ie frame-wise action labels, as inputs for anticipation. We experiment with both action labels and visual features. 

For Breakfast (Table~\ref{tab:SOA_dense_breakfast_50salad}, left), when using GT frame labels, we outperform the others, for shorter prediction horizons. For 50Salads (Table~\ref{tab:SOA_dense_breakfast_50salad}, right), we outperform the state of the art for the observed 20\%, and our predictions are more accurate on long-range anticipation (Pred. 50\%). We outperform~\cite{abu2018will} when we use visual features as input (B Features (Fisher)). When using the segmentations (from~\cite{richard2017weakly}, which has a frame-wise temporal segmentation accuracy of 36.8\% and 42.9\% for the observed 20\% and 30\% of video respectively), we are comparable to state of the art~\cite{Ke_2019_CVPR}. We further merge visual features with action labels for dense anticipation. With Fisher vectors and the frame labels obtained from \cite{richard2017weakly}, we observe a huge performance increase in performance compared to only using the frame labels (up to +7\%) in Breakfast. In 50Salads, this increase is not significant nor consistent. This may be due to the better performing segmentation algorithm on 50Salads (frame-wise accuracy of 66.8\% and 66.7\% for 20\% and 30\% observed respectively). We observe further improvements on Breakfast once we substitute Fisher vectors with I3D features and  segmentations from our own framework (I3D + ours seg.). Similar to next action anticipation, performance drops when using only visual features as input (I3D is better than Fisher vectors). When using I3D features and the frame label outputs of our segmentation method, we obtain our model's best performance, with a slight increase over using only frame label outputs.

\subsection{How much spanning past is necessary?}\label{sec:spanning_past_inf}
\begin{figure}[t]
\centering 
\includegraphics[width=0.7\columnwidth]{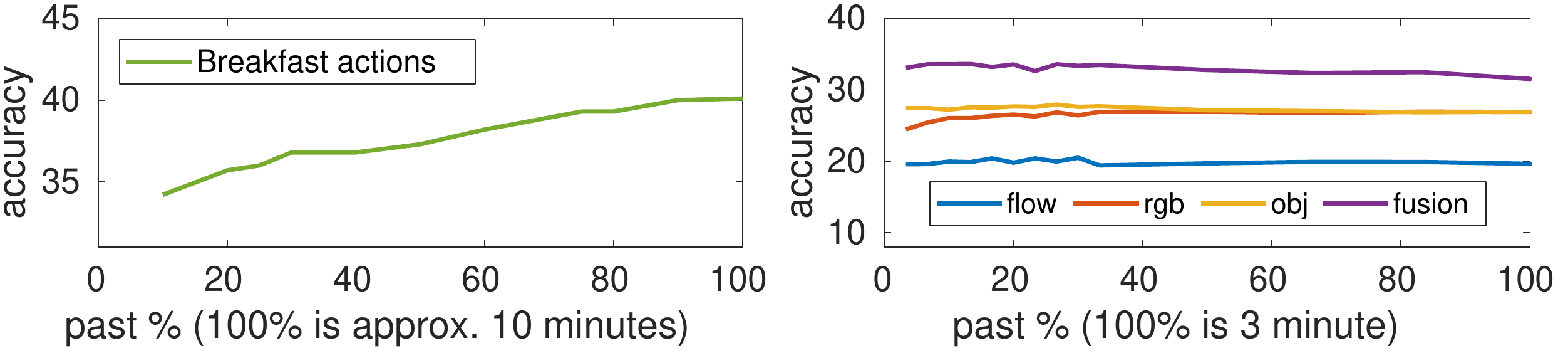}
\caption{Effect of spanning scope on instructional vs.\ daily activities. For EPIC  we report Top-5 Acc. on the validation set with rgb, flow and object features and late fusion.}
\label{fig:pastImportance}
\end{figure}

We vary the duration of spanning snippets (Eq.~\ref{eqn:spanningsnippet}) with start time $i$ as fractions of the current time $t$; $i\!=\!0$ corresponds to the full sequence, \ie 100\% of the spanning past, while $i\!=\!t$ corresponds to none, \ie using only recent snippets since the end points $j$ remain fixed at $t$. Using the entire past is best for Breakfast (Fig.~\ref{fig:pastImportance} left). Interestingly, this effect is not observed on EPIC (Fig.~\ref{fig:pastImportance} right). Though we see a small gain by 1.2\% until 40\% past for the appearance features (rgb), beyond this, performance saturates. We believe this has to do with the fine granularity of labels in EPIC; given that the median action duration is only 1.9s, one could observe as many as 16 actions in 30 seconds. Given that the dataset has only 28.5K samples split over 2513 action classes, we speculate that the model cannot learn all the variants of long-range relationships beyond 30 seconds. Therefore, increasing the scope of the spanning past does not further increase the performance. Based on experiments on the validation set, we set the spanning scope to 6 seconds for EPIC for the rest of the paper.   

\subsection{Recognition and Anticipation on Daily Activities - EPIC}\label{sec:daily_act}

The \textbf{anticipation} task of EPIC requires anticipating the future action $\tau_{\alpha}\!=\!1$s before it starts. For fair comparison to the state of the art~\cite{furnari2019rulstm} (denoted by ``RU''), 
we directly use features (appearance, motion and object) provided by the authors. We train our model separately for each feature modality with the same hyper-parameters and fuse predictions from the different modalities by voting. Note that for experiments on this dataset we do not use the entire past for computing our spanning snippet features (see Section~\ref{sec:spanning_past_inf}).  
Results on hold-out test data of EPIC are given in Table \ref{tab:epic_ex_test} for seen kitchens (S1) with the same environments as the training data and unseen kitchens (S2) of held out environments.
We outperform state of the art, RU~\cite{furnari2019rulstm}, in the Top-1 and Top-5 action accuracies by 2\% and 2.7\% on S1 and by 1.8\% and 2.3\% on S2 using the same features suggesting superior temporal reasoning abilities of our model. When we add verb and noun classification to our model as auxiliary tasks to help with anticipation, ``ours v+n'', our performance improves for action and especially for noun and verb scores.  For challenge results  see supplementary. 

For \textbf{recognition}, we classify pre-trimmed action segments. We adjust the scope of our spanning and recent snippets according to the action start and end times $t_s$ and $t_e$. Spanning  features are computed on a range of $[t_s - 6, t_e + 6]$; the first recent snippet scope is fixed to $[t_s,t_e]$ and the rest to $[t_s-1,t_e+1], [t_s-2,t_e+2]$ and $[t_s-3,t_e+3]$. Remaining hyper-parameters are kept the same. In Table~\ref{tab:epic_ex_test}, we compare to state of the art; we outperform all other  methods including SlowFast networks with audio data~\cite{xiao2020audiovisual} (+5.4\% on S1, +2.2\% on S2 for Top-1) and LFB~\cite{lfb2019}, which also uses non-local blocks (+8.6\% on S1, +5\% on S2 for Top-1) and RU~\cite{furnari2019rulstm} by approximately +7\% on both S1 and S2. Together with the anticipation results we conclude that our method generalizes to both anticipation and recognition tasks and is able to achieve state-of-the-art results on both, while~\cite{furnari2019rulstm} performs very well on anticipation but poorly on recognition.

\begin{table}[t]
\centering
\resizebox{\columnwidth}{!}{ 
\setlength{\tabcolsep}{7pt}
\begin{tabular}{|clccc|ccc|ccc|ccc|}
\hline
& \multicolumn{7}{c|}{\textbf{Action Anticipation}} & \multicolumn{6}{c|}{\textbf{Action Recognition}} \\ \hline
& \multicolumn{4}{c|}{Top-1 Accuracy\%} & \multicolumn{3}{c|}{Top-5 Accuracy\%} & \multicolumn{3}{c|}{Top-1 Accuracy\%} & \multicolumn{3}{c|}{Top-5 Accuracy\%} \\ \hline
& & Verb & Noun & Action & Verb & Noun & Action & Verb & Noun & Action & Verb & Noun & Action \\\hline\rowcolor{red!10}
\multirow{7}{*} 
&\cite{miech2019leveraging} & 30.7 & 16.5 & 9.7 & 76.2 & 42.7 & 25.4 & - & - & - & - & - & - \\\rowcolor{red!10}
& TSN~\cite{damen2018scaling} & 31.8 & 16.2 & 6.0 & 76.6 & 42.2 & 28.2 & 48.2 & 36.7 & 20.5 & 84.1 & 62.3 & 39.8 \\\rowcolor{red!10}
& RU~\cite{furnari2019rulstm} & {33.0} & {22.8} & 14.4 & {79.6} & {50.9} & 33.7 & 56.9 & 43.1 & 33.1 & 85.7 & 67.1 & 55.3 \\\rowcolor{red!10}
\textbf{S1}& LFB~\cite{lfb2019} & - & - & - & - & - & - & 60.0 & 45.0 & 32.7 & 88.4 & {71.8} & 55.3 \\\rowcolor{red!10} 
&\cite{xiao2020audiovisual} & - & - & - & - & - & - & {65.7} & 46.4 & 35.9 & 89.5 & 71.7 & 57.8 \\\rowcolor{red!10}
& ours & 31.4 & 22.6 & {16.4} & 75.2 & 47.2 & \textbf{36.4} & 63.2 & {49.5} & {41.3} & 87.3 & 70.0 & {63.5} \\\rowcolor{red!10}
& ours v+n & \textbf{37.9} & \textbf{24.1} & \textbf{16.6} & \textbf{79.7} & \textbf{54.0} & 36.1 & \textbf{66.7} & \textbf{49.6} & \textbf{41.6} & \textbf{90.1} & \textbf{77.0} & \textbf{64.1} \\\hline\rowcolor{blue!10}
\multirow{7}{*} 
&\cite{miech2019leveraging} & 28.4 & 12.4 & 7.2 & {69.8} & 32.2 & 19.3 & - & - & - & - & - & - \\\rowcolor{blue!10}
& TSN~\cite{damen2018scaling} & 25.3 & 10.4 & 2.4 & 68.3 & 29.5 & 6.6 & 39.4 & 22.7 & 10.9 & 74.3 & 45.7 & 25.3 \\\rowcolor{blue!10}
& RU~\cite{furnari2019rulstm} & 27.0 & 15.2 & 8.2 & 69.6 & {34.4} & 21.1 & 43.7 & 26.8 & 19.5 & 73.3 & 48.3 & 37.2 \\\rowcolor{blue!10}
\textbf{S2}& LFB~\cite{lfb2019} & - & - & - & - & - & - & 50.9 & 31.5 & 21.2 & 77.6 & 57.8 & 39.4 \\\rowcolor{blue!10} 
&\cite{xiao2020audiovisual} & - & - & - & - & - & - & \textbf{55.8} & {32.7} & 24.0 & \textbf{81.7} & {58.9} & 43.2 \\\rowcolor{blue!10}
& ours & {27.5} & \textbf{16.6} & {10.0} & 66.8 & 32.8 & {23.4} & 52.0 & 31.5 & {26.2} & 76.8 & 52.7 & {45.7} \\\rowcolor{blue!10}
& ours v+n & \textbf{29.5} & 16.5 & \textbf{10.1} & \textbf{70.1} & \textbf{37.8} & \textbf{23.4} & 54.6 & \textbf{33.5} & \textbf{27.0} & 80.4 & \textbf{61.0} & \textbf{46.4}\\ \hline
\end{tabular}}
\caption{Action anticipation and recognition on EPIC tests sets S1 and S2} 
\label{tab:epic_ex_test}
\end{table}

\subsection{Temporal Video Segmentation}\label{sec:video_seg}
We compare our performance against the state of the art, MS-TCN (I3D)~\cite{farha2019ms}, in Table~\ref{video_segmentation} on Breakfast. We test our model with 2s and 5s windows. We report the frame-wise accuracy (Acc), segment-wise edit distance (Edit) and F1 scores at overlapping thresholds of 10\%, 25\% and 50\%. In the example sequences, in the F1 scores and edit distances in Table~\ref{video_segmentation}, we observe more fragmentation in our segmentation for 2s than for 5s. However, for 2s, our model produces better accuracies, as the 5s windows are smoothing the predictions at action boundaries. Additionally we provide our model's upper bound, ``ours I3D GT.seg.'', for which we classify GT action segments instead of sliding windows. The results indicate that there is room for improvement, which we leave as future work. We show that we are able to easily adjust our method from its main application and already get close to the state of the art with slight modifications.

 \begin{table}[t]
\centering 
\begin{minipage}[b]{0.45\linewidth}
\centering
\resizebox{\columnwidth}{!}{ 
\setlength{\tabcolsep}{2pt}
\begin{tabular}{|llllll|}
\hline
& \multicolumn{3}{l}{F1@\{10, 25, 50\}}   & Edit & Acc.  \\ \hline
MS-TCN (I3D) ~\cite{farha2019ms} & 52.6 & \multicolumn{1}{l}{48.1} & 37.9 & \textbf{61.7} & \textbf{66.3} \\ \hline
ours I3D 2s          & 52.3 & 46.5                      & 34.8 & 51.3 & 65.3 \\
ours I3D 5s          & \textbf{59.2} & \textbf{53.9} & \textbf{39.5} & 54.5 & 64.5 \\\hline
ours I3D GT.seg.     & -    & -                         & -    & -    & \textbf{75.9} \\ \hline
\end{tabular} } 
\end{minipage} 
\begin{minipage}[b]{0.5\linewidth}
\centering
\includegraphics[width=0.99\columnwidth]{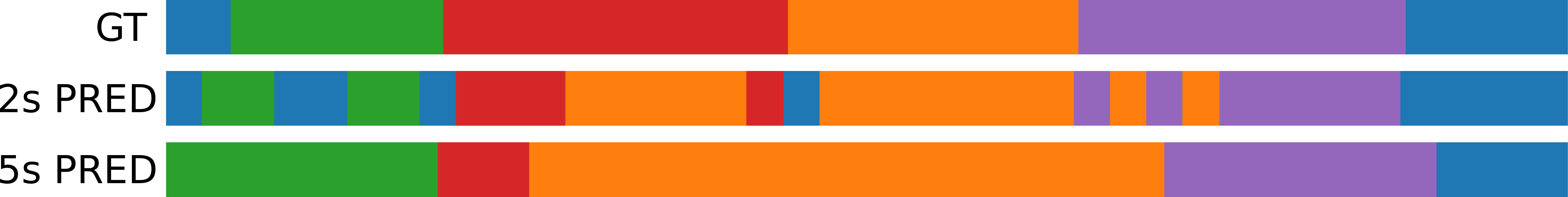}\\
\end{minipage} 
\caption{Exemplary segmentation and comparisons on Breakfast. }
\label{video_segmentation}
\end{table}

\section{Discussion \& Conclusion}

This paper presented a temporal aggregate model for long-range video understanding. Our method computes recent and spanning representations pooled from snippets of video that are related via coupled attention mechanisms. Validating on three complex activity datasets, we show that temporal aggregates are either comparable or outperform the state of the art on three video understanding tasks: action anticipation, recognition and temporal video segmentation.

In developing our framework, we faced questions regarding temporal extent, scaling, and level of semantic abstraction.  Our experiments show that max-pooling is a simple and efficient yet effective way of representing video snippets; this is the case even for snippets as long as two minutes. For learning temporal relationships in long video, attention mechanisms relating the present to long range context can successfully model and anticipate upcoming actions. The extent of context that is beneficial, however, may depend on the nature of activity (instructional vs.~daily) and label granularity (coarse vs.~fine) of the dataset.

We found significant advantages to using ensembles of multiple scales, both in recent and spanning snippets.  Our aggregates model is flexible and can take as input either visual features or frame-wise action labels.  We achieve competitive performance with either form of input, though our experiments confirm that higher levels of abstraction such as labels are more preferable for anticipation.  Nevertheless, there is still a large gap between what can be anticipated with inputs from current segmentation algorithms in comparison to ground truth labels, leaving room for segmentation algorithms to improve.\\ 

\noindent{\textbf{Acknowledgments} This work was funded partly by the German Research Foundation (DFG) YA 447/2-1 and GA 1927/4-1 (FOR2535 Anticipating Human Behavior) and partly by National Research Foundation Singapore under its NRF Fellowship Programme [NRF-NRFFAI1-2019-0001] and Singapore Ministry of Education (MOE) Academic Research Fund Tier 1 T1251RES1819.}
\clearpage 
\bibliographystyle{splncs04}
\bibliography{egbib}
\end{document}


\pagestyle{headings}
\mainmatter
\def\ECCVSubNumber{2515} 

\title{Supplementary: Temporal Aggregate Representations\\
for Long-Range Video Understanding} 

\titlerunning{Temporal Aggregate Representations for Long-Range Video Understanding}
%
\author{Fadime Sener\inst{1,2} \and
Dipika Singhania\inst{2}\and
Angela Yao\inst{2}}
\authorrunning{Sener et al.}
\institute{University of Bonn, Germany \and
National University of Singapore \\
\email{\{sener@cs.uni-bonn.de\},\{dipika16,ayao\}@comp.nus.edu.sg}}
\maketitle

\section{More on Datasets and Features}
We provide more statistics about the datasets used in our paper to show a broader comparison about their scale and label granularity. \\
 
\noindent\textbf{The Breakfast Actions dataset \cite{kuehne2014language}} contains 1712 videos of 10 high level tasks like ``making coffee'', ``making tea'' and so on. There are in total 48 different actions, such as ``pouring water'' or ``stirring coffee'', with on average 6 actions per video. The average duration of the videos is 2.3 minutes. There are 4 splits and we report our results averaged over them. We use two types of frame-wise features: Fisher vectors computed as in \cite{abu2018will} and I3D features~\cite{carreira2017quo}. 

\noindent\textbf{The 50Salads dataset~\cite{stein2013combining}} includes 50 videos and 17 different actions for a single task, namely making mixed salads. When training on this dataset, we therefore omit task prediction in our model. On average, 50Salads has 20 actions per video due to repetitions. The average video duration is 6.4 minutes. There are 5 splits, and we again average our results over them. We represent the frames using Fisher vectors as in \cite{abu2018will}.

\noindent\textbf{The EPIC-Kitchens dataset~\cite{damen2018scaling}} is a large first-person video dataset which contains 432 sequences and 39,594 action segments recorded by participants performing non-scripted daily activities in their kitchen. The average duration of the videos is 7.6 minutes ranging from 1 minute to 55 minutes. An action is defined as a combination of a verb and a noun, e.g. ``boil milk``. There are in total 125 verbs, 351 nouns and 2513 actions. The dataset provides a training and test set which contains 272 and 160 videos, respectively. The test set is divided into two splits: Seen Kitchens (S1) where sequences from the same environment are in the training data, and Unseen Kitchens (S2) where complete sequences of some participants are held out for testing. The labels for the test set are not shared, as there is an action anticipation challenge\footnote{ \texttt{\url{https://competitions.codalab.org/competitions/20071}}} and action recognition challenge\footnote{ \texttt{\url{https://competitions.codalab.org/competitions/20115}}}. We use the RGB, optical flow and object-based features provided by Furnari and Farinella~\etal~\cite{furnari2019rulstm}. The minimum and maximum snippet durations, over which we apply pooling, are 0.4s and 115.3s for 50Salads, 0.1s and 64.5s for Breakfast, and 1.2s and 3.0s for EPIC.

\noindent\textbf{Implementation Details : } We train our model using the Adam optimizer~\cite{kingma2014adam} with batch size 10, learning rate $10^{-4}$ and dropout rate 0.3. We train for 25 epochs and decrease the learning rate by a factor of 10 every $10^{\text{th}}$ epoch.   We use 1024 dimensions for all non-classification linear layers for the Breakfast Actions and 50Salads datasets and 512 dimensions for the EPIC-Kitchens dataset. The LSTMs in dense anticipation have one layer and 512 hidden units. We use intervals of $20$  seconds for Breakfast and 50Salads for discretizing the durations in dense anticipation.

\section{Model Validation} 
\begin{table}[!]
\centering
\resizebox{\columnwidth}{!}{ 
\setlength{\tabcolsep}{6pt}
\begin{tabular}{|llllllllllll|}
\hline
& cereal & coffee & f.egg & juice & milk & panc. & salat & sand. & s.egg & tea & mean\scriptsize{$\pm$std}\\\hline 
TM &\cellcolor[HTML]{4A9BD0}77.8 &\cellcolor[HTML]{C7E0F0}50.8 &\cellcolor[HTML]{ACD1EA}57.2 &\cellcolor[HTML]{ACD1EA}57.2 &\cellcolor[HTML]{EBF4FA}40.1 &\cellcolor[HTML]{ECF4FA}39.6 &\cellcolor[HTML]{A9CFE9}57.9 &\cellcolor[HTML]{C0DCEF}52.4 &\cellcolor[HTML]{A2CCE7}59.4 &\cellcolor[HTML]{B9D8ED}54.2 &\cellcolor[HTML]{B7D7EC}54.6\scriptsize{$\pm$10.8} \\ \hline
LUT &\cellcolor[HTML]{ABD0E9}57.5 &\cellcolor[HTML]{A0CAE6}\textbf{59.9} &\cellcolor[HTML]{B0D3EB}56.2 &\cellcolor[HTML]{A5CDE8}58.8 &\cellcolor[HTML]{B1D4EB}56.1 &\cellcolor[HTML]{ACD1E9}57.3 &\cellcolor[HTML]{B5D6EC}55.1 &\cellcolor[HTML]{CBE2F2}49.6 &\cellcolor[HTML]{9AC7E5}\textbf{61.2} &\cellcolor[HTML]{9FCAE6}60.1 &\cellcolor[HTML]{ACD1EA}57.2\scriptsize{$\pm$3.1} \\ \hline
LSTM &\cellcolor[HTML]{4095CD}\textbf{79.8} &\cellcolor[HTML]{D5E8F4}47.2 &\cellcolor[HTML]{BEDBEE}52.9 &\cellcolor[HTML]{9AC7E5}61.2 &\cellcolor[HTML]{64A9D7}72.7 &\cellcolor[HTML]{5EA6D5}\textbf{73.9} &\cellcolor[HTML]{8CBFE1}\textbf{64.3} &\cellcolor[HTML]{D6E8F4}46.9 &\cellcolor[HTML]{9DC9E6}60.5 &\cellcolor[HTML]{77B4DC}\textbf{68.7} &\cellcolor[HTML]{93C3E3}62.8\scriptsize{$\pm$11.3} \\ \hline
ours &\cellcolor[HTML]{72B1DA}69.8 &\cellcolor[HTML]{B7D7EC}54.7 &\cellcolor[HTML]{94C4E3}\textbf{62.5} &\cellcolor[HTML]{85BCDF}\textbf{65.7} &\cellcolor[HTML]{63A8D6}\textbf{72.9} &\cellcolor[HTML]{83BADF}66.2 &\cellcolor[HTML]{8FC1E2}63.6 &\cellcolor[HTML]{8BBFE1}\textbf{64.6} &\cellcolor[HTML]{A8CFE9}58.0 &\cellcolor[HTML]{8DC0E1}64.1 &\cellcolor[HTML]{8CC0E1}\textbf{64.2}\scriptsize{$\pm$5.2} \\ \hline
\end{tabular}}
\caption{
Model validation using GT labels for next action anticipation on the Breakfast Actions, presented are accuracies. 
We compare transition matrices (TM), lookup tables (LUT), LSTMs, and our temporal aggregates model (without complex activity prediction).
}
\label{tab:baselineComparison}
\end{table}

For validating our method's capabilities in modelling sequences, we make baseline comparisons.
The simplest approach for solving the next action anticipation task is using a transition matrix (TM)~\cite{miech2019leveraging}, which encodes the transition from one action to the next. A more sophisticated solution is building a lookup table (LUT) of varying length sequences which allows encoding the context in a more explicit manner. The problem with LUTs is that their completeness depends on the coverage of the training data, and they rapidly grow with the number of actions. So far, for next step prediction, RNNs achieve good performance \cite{abu2018will}, as they learn modelling the sequences. 

For our baseline comparisons, instead of frame features, we use the frame-level ground truth labels as input to our model. We compute the TM, LUT and RNN on the ground truth segment-level labels. In Table \ref{tab:baselineComparison} we present comparisons on the Breakfast Actions for the next action anticipation per complex activity. Overall, transition matrices provide the worst results. LUTs improve the results, as they incorporate more contextual information. Both the RNN and our method outperform the other alternatives, while our method still performs better than the RNN on average. However, applying RNNs requires parsing the past into action sequences~\cite{abu2018will}, which turns the problem into separate segmentation and prediction phases. Our model, on the other hand, can be trained end-to-end, and can represent the long-range observations good enough to outperform RNNs.  We show that our model is doing better than simply learning pairwise statistics of the dataset.

\begin{figure}[t!]
\centering 
\includegraphics[width=0.55\columnwidth]{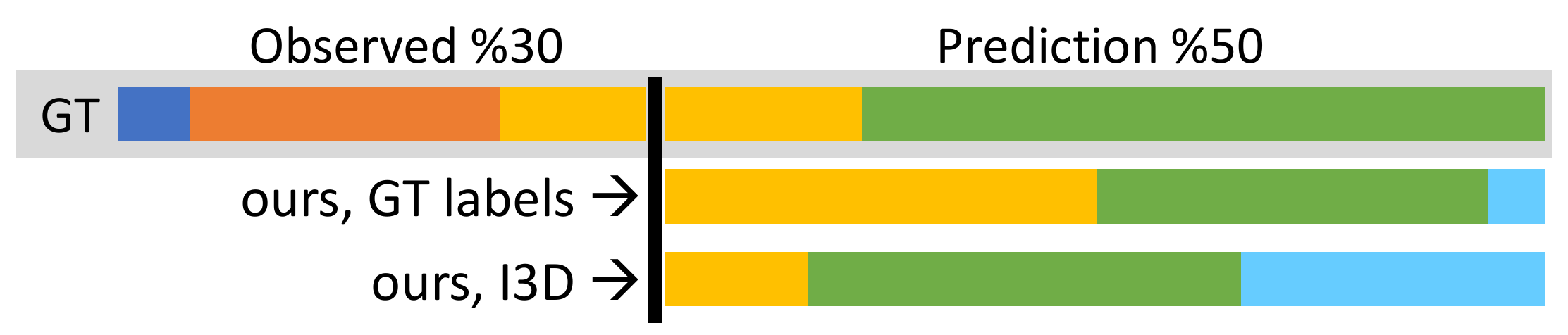}  
\includegraphics[width=0.55\columnwidth]{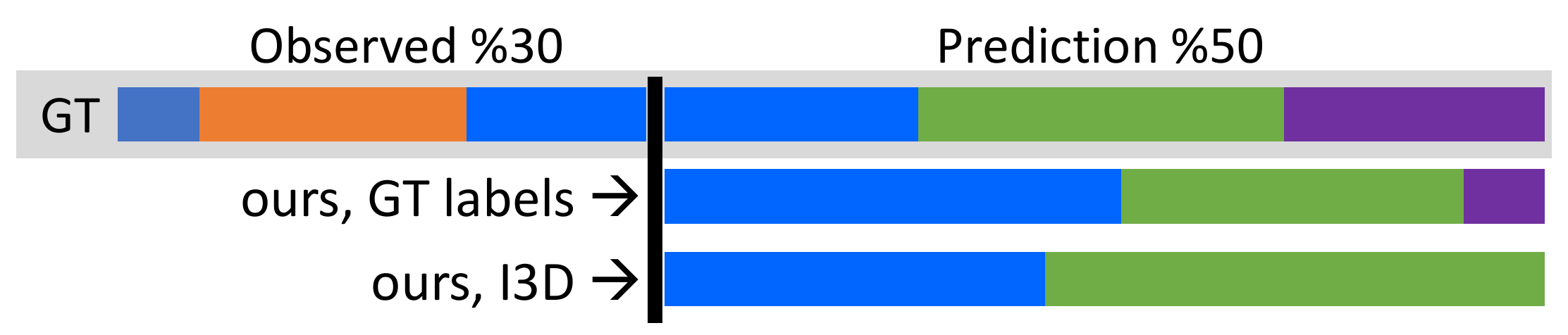}
\includegraphics[width=0.55\columnwidth]{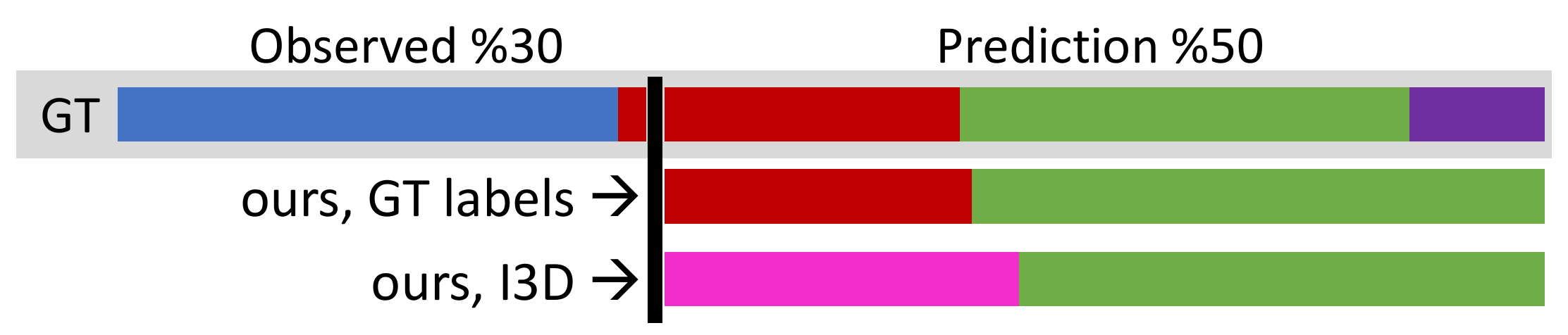}
\includegraphics[width=0.55\columnwidth]{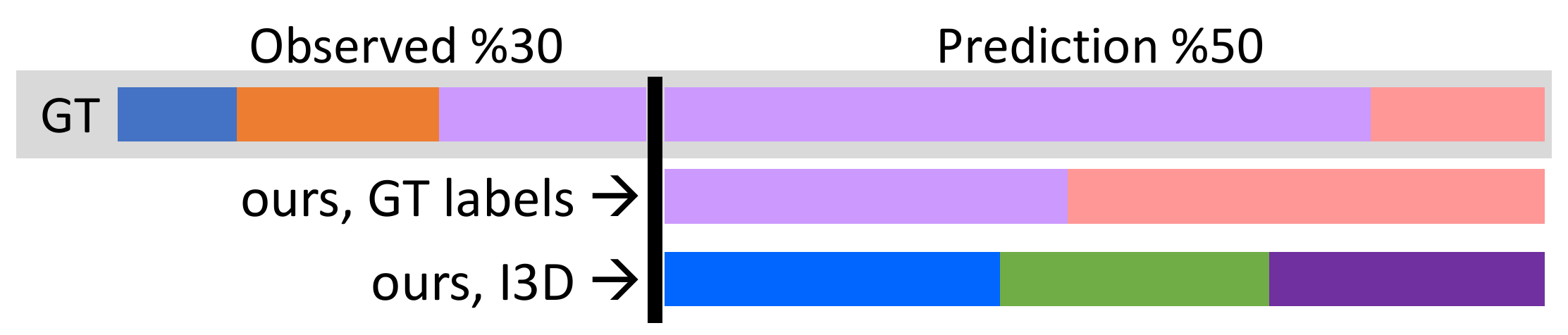}
\includegraphics[width=0.55\columnwidth]{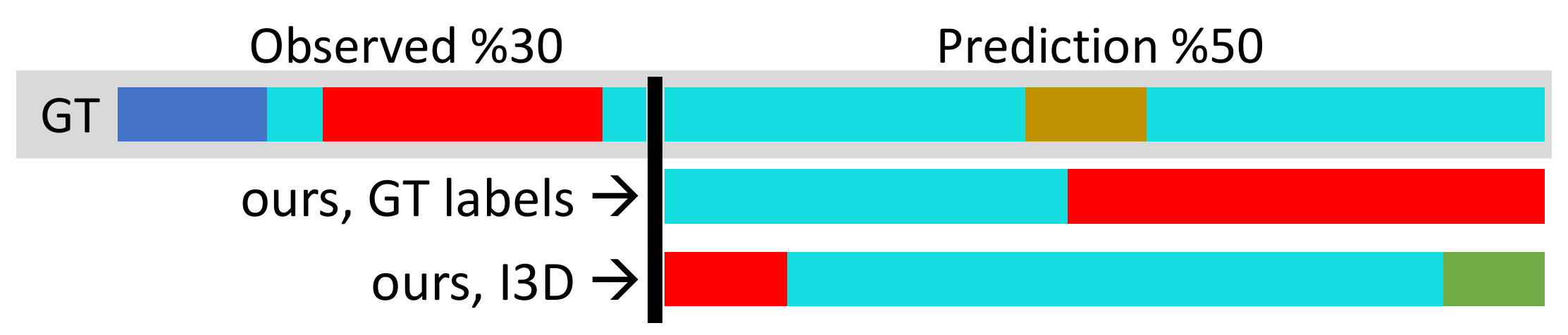}
\includegraphics[width=0.55\columnwidth]{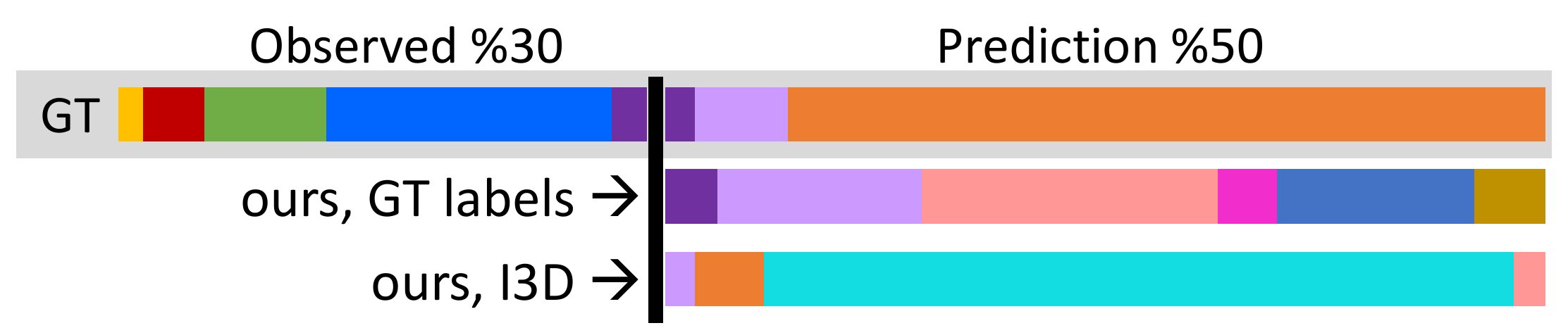}
\caption{Qualitative results for dense anticipation on Breakfast Actions dataset when using the GT labels and I3D features. Best viewed in color.}
\label{fig:dense_vis} 
\end{figure}

\begin{figure*}[t!]
\centering 
\includegraphics[width=0.9\textwidth]{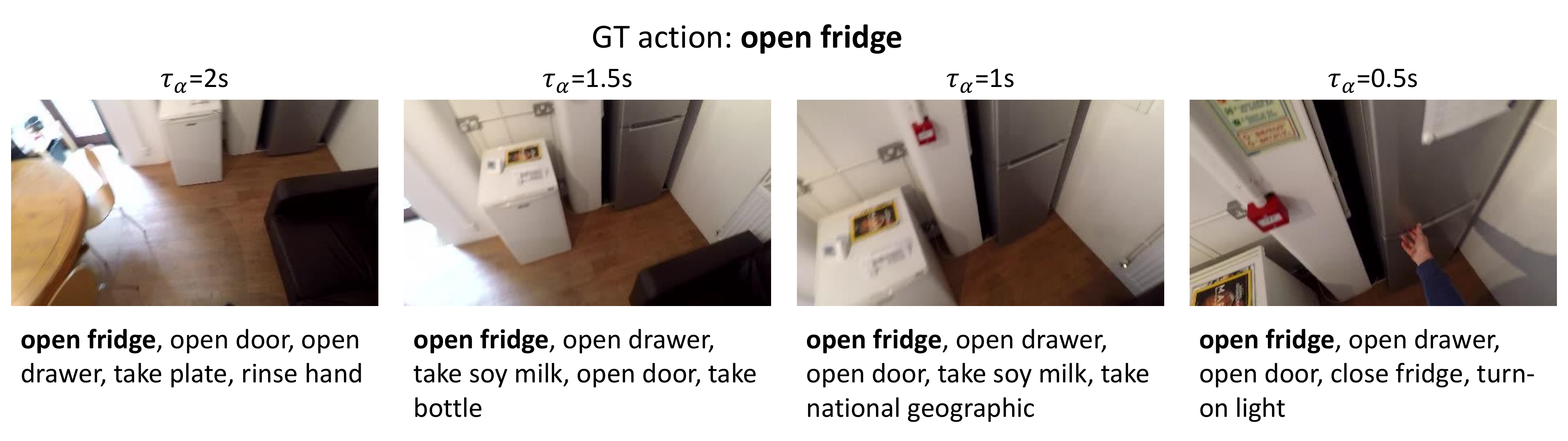}
\includegraphics[width=0.9\textwidth]{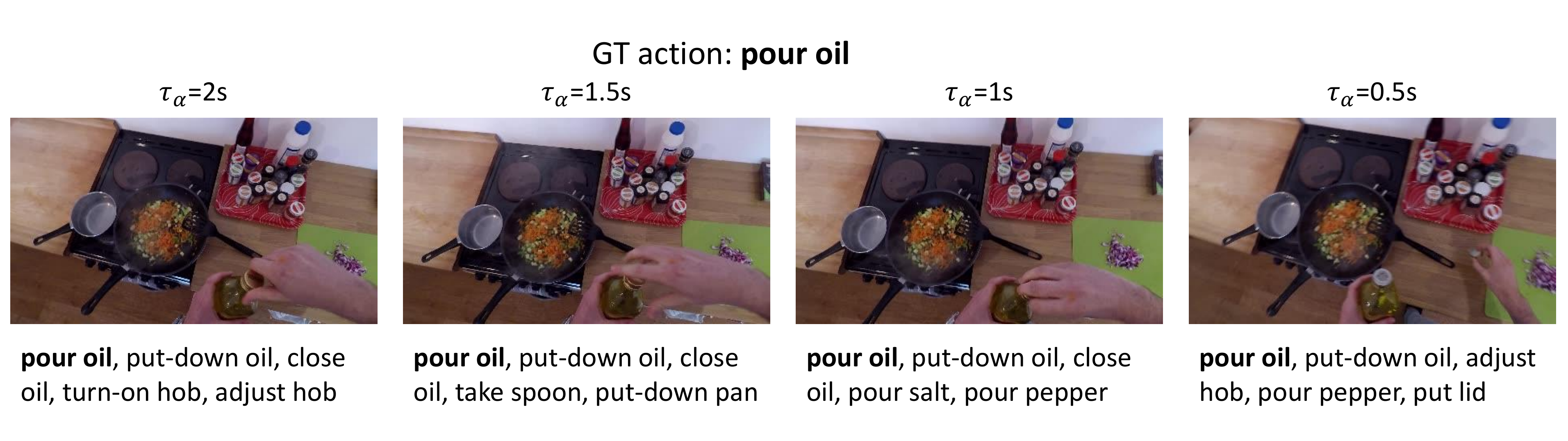} 
\includegraphics[width=0.9\textwidth]{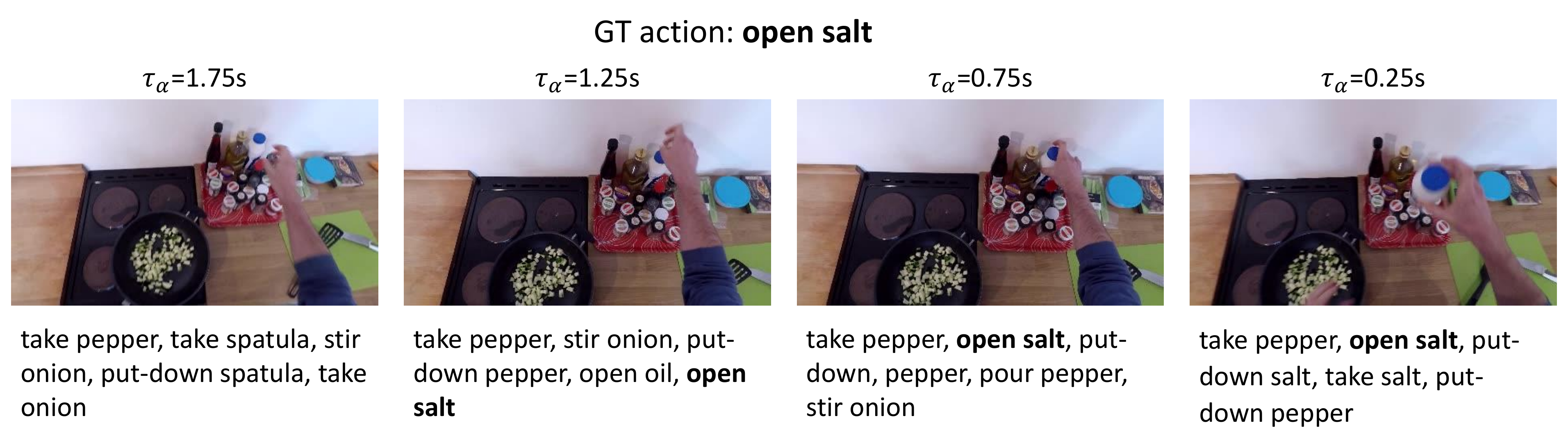} 
\includegraphics[width=0.9\textwidth]{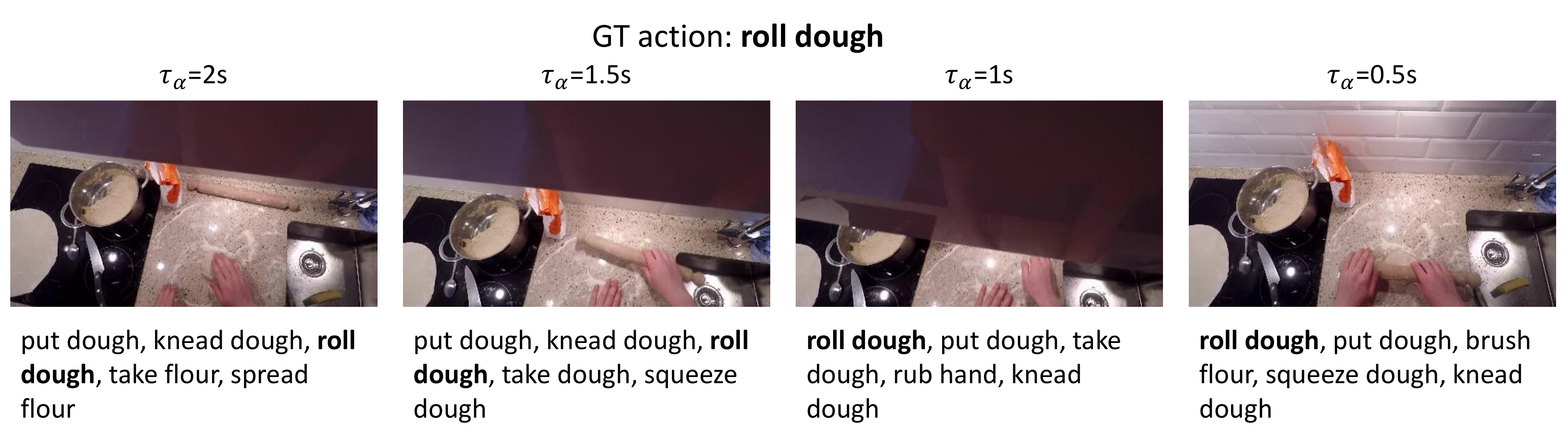} 
\caption{Exemplary qualitative results for next action anticipation on EPIC-Kitchens dataset, showing the success of our method. We list our Top-5 predictions at different anticipation times, $\tau_{\alpha}$. The closer we are the better are our model's predictions. Best viewed in color.}
\label{fig:epic_vis} 
\end{figure*} 

\begin{figure*}[t!]
\centering 
\includegraphics[width=0.9\textwidth]{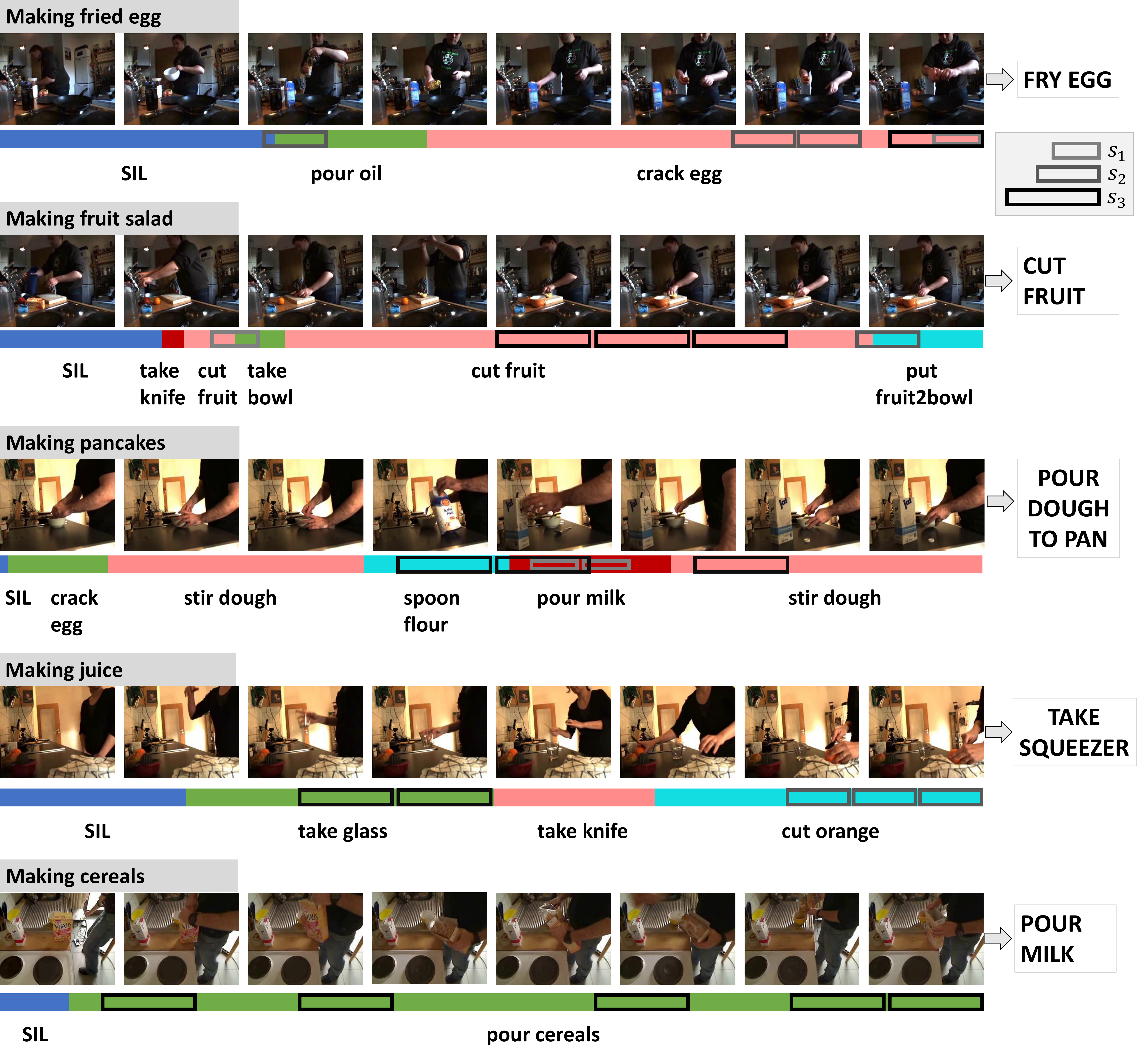} 
\caption{Attention visualization on the Breakfast Actions dataset for next action anticipation. Rectangles are the top 5 five spanning snippets (at different granularities where K = {10,15,20}), weighted highest by the attention mechanism in the Non-Local Blocks (NLB). Best viewed in color.}
\label{fig:atten_vis} 
\end{figure*}

\section{Visual Results}
In Fig.~\ref{fig:dense_vis}, we provide qualitative results from our method for dense anticipation on the Breakfast Actions dataset. We show our method's predictions after observing 30\% of the video. We compare our results when we use the GT labels and I3D features as input. 
 
In Fig.~\ref{fig:epic_vis}, we present qualitative results from our method for next action anticipation on the EPIC-Kitchens dataset for multiple anticipation times $\tau_{\alpha}$ between $0.25$ and $2$ seconds. We show examples where our method is certain about the next action for all different times. We also show examples where our method's prediction gets more accurate when the anticipation time is closer. 
 
In Fig.~\ref{fig:atten_vis}, we present some visualizations of regions attended by our non-local blocks. We show the five highest weighted spanning snippets (at different granularities). Our model attends different regions over the videos, for example for predicting 'fry egg' when making fried eggs, it attends regions both when pouring oil and cracking eggs. Pouring oil is an important long-range past action for frying eggs.  Our method can encode long video durations while attending to salient snippets.

\section{Action Anticipation on EPIC-Kitchens}
Furnari and Farinella~\cite{furnari2019rulstm} reports prediction results at multiple anticipation times ($\tau_{\alpha}$) between $0.25$s and $2$s on EPIC. We compare in Table~\ref{tab:anticipation_ek} on the validation set and note that our prediction scores are better than~\cite{furnari2019rulstm} for all time points.  Our improvements are greater when the anticipation time decreases.  

\begin{table}[t]
\centering
\resizebox{\columnwidth}{!}{ 
\setlength{\tabcolsep}{13pt}
\begin{tabular}{|rrrrrrrrr|}
\hline
\multicolumn{1}{|c}{} & \multicolumn{8}{c|}{Top-5 ACTION Accuracy\%} \\ \hline
 $\tau_a$ & 2s & 1.75s & 1.5s & 1.25s & 1.0s & 0.75s & 0.5s & 0.25s \\ \hline 
RU~\cite{furnari2019rulstm} & 29.4 & 30.7 & 32.2 & 33.4 & 35.3 & 36.3 & 37.4 & 39.0 \\ 
\textbf{ours} & \textbf{30.9} & \textbf{31.8} & \textbf{33.7} & \textbf{35.1} & \textbf{36.4} & \textbf{37.2} & \textbf{39.5} & \textbf{41.3} \\ \hline
\end{tabular} }
\caption{Action anticipation on EPIC validation set at different anticipation times.}
\label{tab:anticipation_ek}
\end{table}

We report our results for hold-out test data on EPIC-Kitchens Egocentric Action Anticipation Challenge (2020) in Table \ref{tab:epic_ex_test} for seen kitchens (S1) with the same environments as in the training data and unseen kitchens (S2) of held out environments.   The official ranking on the challenge is based on the Top-1 action accuracy. Our submission (Team ``NUS\_CVML'') is ranked first on S1 and third on S2 sets. We refer the reader to  EPIC-Kitchens 2020 Challenges Report~\cite{epicreport20}  for  details on the competing methods.   

\begin{table*}[!t]
\centering
\resizebox{\textwidth}{!}{{}
\setlength{\tabcolsep}{2pt}
\begin{tabular}{|clccc|ccc|ccc|ccc|}
\hline
& \multicolumn{4}{c|}{Top-1 Accuracy\%} & \multicolumn{3}{c|}{Top-5 Accuracy\%} & \multicolumn{3}{c|}{ Precision (\%)} & \multicolumn{3}{c|}{Recall (\%)} \\ \hline
& & Verb & Noun & Action & Verb & Noun & Action & Verb & Noun & Action & Verb & Noun & Action \\\hline\rowcolor{red!10}
\multirow{7}{*} 
 & \textbf{1st} (S1) & 37.87 &	24.10 &	\textbf{16.64} &	79.74 &	53.98 &	36.06 &	36.41 &	25.20 & 9.64 &	15.67 &	22.01 &	10.05 \\\hline\rowcolor{blue!10}
\multirow{7}{*} 
 & \textbf{3rd} (S2) & 29.50 & 16.52 & \textbf{10.04} & 70.13 & 37.83 & 	23.42 &	20.43 &	12.95 & 4.92 &	8.03 &	12.84 &	6.26
\\ \hline
\end{tabular}}
\caption{Action anticipation on EPIC tests sets, seen (S1) and unseen (S2)}
\label{tab:epic_ex_test}
\end{table*} 
 
\begin{table*}[!t]
\centering
\resizebox{\textwidth}{!}{{}
\setlength{\tabcolsep}{2pt}
\begin{tabular}{|clccc|ccc|ccc|ccc|}
\hline
& \multicolumn{4}{c|}{Top-1 Accuracy\%} & \multicolumn{3}{c|}{Top-5 Accuracy\%} & \multicolumn{3}{c|}{ Precision (\%)} & \multicolumn{3}{c|}{Recall (\%)} \\ \hline
& & Verb & Noun & Action & Verb & Noun & Action & Verb & Noun & Action & Verb & Noun & Action \\\hline\rowcolor{red!10}
\multirow{7}{*} 
 & \textbf{2nd} (S1) &  66.56 &	49.60 & \textbf{41.59} & 90.10 & 77.03 & 64.11 & 59.43 & 45.62 &	25.37 &	41.65 &	46.25 &	26.98 \\\hline\rowcolor{blue!10}
\multirow{7}{*} 
& \textbf{3rd} (S2)  & 54.56 & 33.46 & \textbf{26.97} &	80.40 &	60.98 &	46.43 &	33.60 &	30.54 &	14.99 &	25.28 &	28.39 &	17.97
\\ \hline
\end{tabular}}
\caption{Action recognition on EPIC tests sets, seen (S1) and unseen (S2)}
\label{tab:epic_ex_test_rec}
\end{table*}

\section{Action Recognition Challenge on EPIC-Kitchens}
We present our results for the EPIC-Kitchens Egocentric Action Recognition Challenge 2020 in Table \ref{tab:epic_ex_test_rec} for S1 and S2. Our team  ``NUS\_CVML''  is ranked second on S1 and third on S2 sets.  Please see EPIC-Kitchens 2020 Challenges Report~\cite{epicreport20}  for further details. 
 
\bibliographystyle{splncs04}
\bibliography{egbib}